\documentclass[8pt,twocolumn]{IEEEtran}

\makeatletter
\def\ifundefined{\@ifundefined}
\makeatother

\usepackage{epsfig}
\usepackage{dsfont}
\usepackage{amsmath}
\usepackage{amssymb} 
\usepackage{graphicx}
\usepackage{amsthm}
\usepackage[normalem]{ulem}
\usepackage{subfigure}
\usepackage{color}
\usepackage{hyperref}
\usepackage{bm}
\usepackage{lineno}
\usepackage{multirow}
\usepackage[ruled,vlined]{algorithm2e}
\usepackage{amsmath}
\usepackage{cite}
\usepackage{lettrine}
\usepackage{diagbox} %For Mac
\usepackage{rotating}

\makeatletter
\let\start@align@nopar\start@align
\let\start@gather@nopar\start@gather
\let\start@multline@nopar\start@multline
\long\def\start@align{\par\start@align@nopar}
\long\def\start@gather{\par\start@gather@nopar}
\long\def\start@multline{\par\start@multline@nopar}
\makeatother
\newcommand{\be}{\begin{equation}}
\newcommand{\ee}{\end{equation}}
\newcommand{\bea}{\begin{eqnarray}}
\newcommand{\eea}{\end{eqnarray}}
\newcommand\ba[1]{\left[ \begin{array}{#1}}
	\def\ea{\end{array}\right]}

\newcommand{\bfi}{\begin{figure}}
	\newcommand{\efi}{\end{figure}}

\newcommand{\C}{\mathbb{C}}

\newcommand{\R}{\mathbb{R}}

\newcommand{\X}{\mathcal{X}}

\newcommand{\x}{\mathbf{x}}
\newcommand{\y}{\mathbf{y}}
\newcommand{\w}{\mathbf{w}}
\newcommand{\inner}[1]{\langle {#1} \rangle}

\providecommand{\abs}[1]{\lvert#1\rvert}
\providecommand{\norm}[1]{\lVert#1\rVert}

\newtheorem{thm}{Theorem}  %   number by Chapter
\newtheorem{pro}{Proposition}

\newtheorem{defn}{Definition}       %   Numbered independently from thm
\newtheorem{corollary}{Corollary}       %   Numbered independently from thm

\begin{document}
	\title{No-Trick (Treat) Kernel Adaptive Filtering using Deterministic Features
		\thanks{This work was supported by the Lifelong Learning Machines program from DARPA/MTO grant FA9453-18-1-0039.}
		\thanks{The authors are with the Computational NeuroEngineering Laboratory, University of Florida, Gainesville, FL 32611 USA  (e-mail: likan@ufl.edu; principe@cnel.ufl.edu).}}
	\author{Kan Li, \IEEEmembership{Member,~IEEE} and Jos\'{e} C. Pr\'{i}ncipe, \IEEEmembership{Fellow,~IEEE}}
	\markboth{}%
	{Li \MakeLowercase{\textit{et al.}}: }
		
	\maketitle
	
\begin{abstract}
Kernel methods form a powerful, versatile, and theoretically-grounded unifying framework to solve nonlinear problems in signal processing and machine learning.	The standard approach relies on the \emph{kernel trick} to perform pairwise evaluations of a kernel function, which leads to scalability issues for large datasets due to its linear and superlinear growth with respect to the training data. A popular approach to tackle this problem, known as random Fourier features (RFFs), samples from a distribution to obtain the data-independent basis of a higher finite-dimensional feature space, where its dot product approximates the kernel function. Recently, deterministic, rather than random construction has been shown to outperform RFFs, by approximating the kernel in the frequency domain using Gaussian quadrature. In this paper, we view the dot product of these explicit mappings not as an approximation, but as an equivalent positive-definite kernel that induces a new finite-dimensional reproducing kernel Hilbert space (RKHS). This opens the door to \emph{no-trick} (NT) online kernel adaptive filtering (KAF) that is scalable and robust. Random features are prone to large variances in performance, especially for smaller dimensions. Here, we focus on deterministic feature-map construction based on polynomial-exact solutions and show their superiority over random constructions. Without loss of generality, we apply this approach to classical adaptive filtering algorithms and validate the methodology to show that deterministic features are faster to generate and outperform state-of-the-art kernel methods based on random Fourier features.
\end{abstract}
	
\section{Introduction}
Kernel methods form a powerful, flexible, and principled framework to solve nonlinear problems in signal processing and machine learning. They have attracted a resurgent interest from competitive performances with deep neural networks in certain tasks \cite{rahimi2007RFF,NIPS2009_3628, Huang2014,Wilson2016,Li2018}.	The standard approach relies on the representer theorem and the \textit{kernel trick} to perform pairwise evaluations of a kernel function, which leads to scalability issues for large datasets due to its linear and superlinear growth with respect to the training data. A popular approach to handling this problem, known as random Fourier features, samples from a distribution to obtain the basis of a lower-dimensional space, where its dot product approximates the kernel evaluation. Employing randomized feature map means that $O(\epsilon^{-2})$ samples are required to achieve an approximation error of at most $\epsilon$ \cite{rahimi2007RFF,Dao2017}. Recently, alternative deterministic, rather than random, construction has been shown to outperform random features, by approximating the kernel in the frequency domain using Gaussian quadrature. It has been shown that deterministic feature maps can be constructed, for any $\gamma > 0$, to achieve error $\epsilon$ with $O(e^{e^\gamma} + \epsilon^{-1/\gamma})$ samples as $\epsilon$ goes to 0 \cite{Dao2017}. In this paper, we adopt this approach and generalize to a class of polynomial-exact solutions for applications in kernel adaptive filtering (KAF). We validate the methods using simulation to show that deterministic features are faster to generate and outperform state-of-the-art kernel approximation methods based on random Fourier features, and require significantly fewer resources to compute than their conventional KAF counterparts.
	
\begin{figure}[t]
	\centering
	\includegraphics[width=0.48\textwidth]{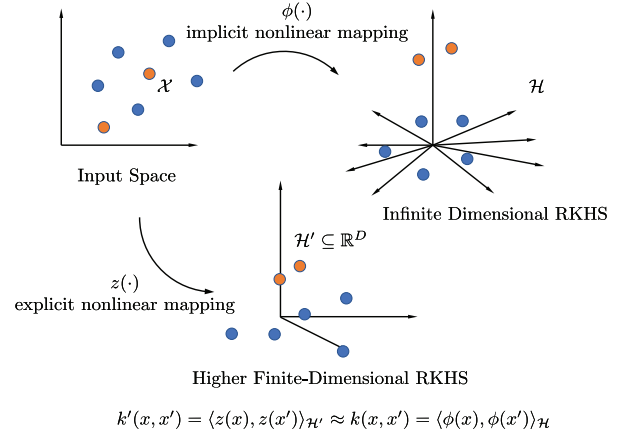}
	\caption{Under mild conditions, explicit nonlinear features define an equivalent reproducing kernel which induces a new higher finite-dimensional RKHS where its dot product value approximates the inner product of a potentially infinite-dimensional RKHS.}
	\label{fig:featurespace}
\end{figure} 
	In the standard kernel approach, points in the input space $\mathbf{x}_i\in\mathcal{X}$ are mapped, using an implicit nonlinear function $\phi(\cdot)$, into a potentially infinite-dimensional inner product space or reproducing kernel Hilbert space (RKHS), denoted by $\mathcal{H}$. The explicit representation is of secondary nature. A real valued similarity function $k:\mathcal{X}\times\mathcal{X}\rightarrow \R$ is defined as
	\begin{equation}
	k(\mathbf{x},\mathbf{x}')=\langle\phi(\mathbf{x}),\phi(\mathbf{x}')\rangle
	\end{equation}
	which is referred to as a reproducing kernel. This presents an elegant solution for classification, clustering, regression, and principal component analysis, since the mapped data points are linearly separable in the potentially infinite-dimensional RKHS, allowing classical linear methods to be applied directly on the data. However, because the actually points (functions) in the function space are inaccessible, kernel methods scale poorly to large datasets. Naive kernel methods operate on the kernel or Gram matrix, whose entries are denoted $G_{i,j}=k(\mathbf{x}_i,\mathbf{x}_j)$, requiring $O(n^2)$ space complexity and $O(n^3)$ computational complexity for many standard operations. For online kernel adaptive filtering algorithms, this represents a rolling sum with linear or superlinear growth. There have been a continual effort to sparsify and reduce the computational load, especially for online KAF \cite{QKLMS, NICE,SNIPGOAL}.
	
An approximation approach was proposed in \cite{rahimi2007RFF}, where the kernel is estimated using the dot product in a higher finite-dimensional space, $\R^D$, (which we will refer to as the feature space), using data-independent feature map $z: \mathcal{X} \rightarrow \R^D$ such that $\langle z(\mathbf{x}),z(\mathbf{x}')\rangle\approx k(\mathbf{x}, \mathbf{x}')$. This approximation enables scalable classical linear filtering techniques to be applied directly on the explicitly transformed data without the computational drawback of the naive kernel methods. This approach has been applied to various kernel adaptive filtering algorithms \cite{Singh12, Qi15, Bouboulis18}. The quality of this approximation, however, is not well understood, as it requires random sampling to approximate the kernel. This is an active research area with many error bounds proposed \cite{rahimi2007RFF,Sutherland2015,Sun2018,pmlr-v97-li19k}. Furthermore, it has been proven that the variant of RFF features used in these existing KAF papers have strictly higher variance for the Gaussian kernel with worse bounds \cite{Sutherland2015}. Here, we adopt the alternative, deterministic framework that provides exact-polynomial solutions such as using Gaussian quadrature \cite{Dao2017} and Taylor series expansion \cite{cotter2011explicit}, which is related to the fast Gauss transform in kernel density estimation \cite{Greengard1991, Yang2003}. This eliminates the undesirable effects of performance variance in random features generation. For example, the finite truncations of the Taylor series of a real-valued function $f(x)$ about the point $x = a$ are all exactly equal to $f(\cdot)$ at $a$. On the other hand, the Fourier series (generalized by the Fourier Transform) is integrated over the entire interval, hence there is generally no such point where all the finite truncations are exact.

Rather than viewing an explicit feature space mapping, such as random Fourier features, as an approximation $\tilde{k}$ to a kernel function $k$, it is more fitting to view it as an equivalent kernel that induces a new RKHS: a nonlinear mapping $z(\cdot)$ that transforms the data from the original input space to a higher finite-dimensional RKHS, $\mathcal{H}'$, where $k'(\x,\x')=\langle z(\x),z(\x')\rangle_{\mathcal{H}'}$ (Fig. \ref{fig:featurespace}). It is easy to show that the mappings induce a positive definite kernel function satisfying Mercer's conditions under the closer properties (where, positive-definite kernels, such as exponentials and polynomials, are closed under addition, multiplication, and scaling). It follows that these kernels are universal: they approximate uniformly an arbitrary continuous target function over any compact domain.

Consequently, performing classical linear filtering in this new RKHS produces nonlinear solutions in the original input space. Unlike conventional kernel methods, we do not need to rely on the \emph{kernel trick}. Defining data-independent basis decouples the data from the projection, allowing explicit mapping and weight vector formulation in the finite-dimensional RKHS. This frees the data from the rolling sum, since the weight vectors can now be consolidated in each iteration.

The rest of the paper is organized as follows. In Section \ref{Sec:RKHS}, deterministic feature constructions are discussed and we propose the concept of explicit feature-map RKHS. Kernel adaptive filtering is reviewed in Section \ref{Sec:KAF}, and no-trick (NT) KAF is presented. Experimental results are shown in Section \ref{Sec:Simulation}. Finally, Section \ref{Sec:Conclusion} concludes this paper.
\section{Explicit Feature-Map Reproducing Kernel}\label{Sec:RKHS}
In this section, we first present the theorem that gave rise to random feature mappings, then introduce polynomial-exact deterministic features, and show that these mappings define a new equivalent reproducing kernel with universal approximation property.

The popular random Fourier features belong to a class of simple, effective methods for scaling up kernel machines that leverages the following underlying principle.
\begin{thm}[Bochner, 1932\cite{Bochner1959}]
	A continuous shift-invariant properly-scaled kernel $k(\mathbf{x},\mathbf{x}')=k(\mathbf{x}-\mathbf{x}'):\R^d\times\R^d\rightarrow\R$, and $k(\x,\x)=1,\forall\x$, is positive definite if and only if $k$ is the Fourier transform of a proper probability distribution.
\end{thm}
The corresponding kernel can then be expressed in terms of its Fourier transform $p(\bm{\omega})$ (a probability distribution) as
\begin{align}
k(\mathbf{x}-\mathbf{x}')&=\int_{\R^d}p(\bm{\omega})e^{j\bm{\omega}^\intercal(\mathbf{x}-\mathbf{x}')}d\bm{\omega}\label{eq:kintergral}\\
&={\rm E}_{\bm{\omega}}\left[e^{j\bm{\omega}^\intercal(\mathbf{x}-\mathbf{x}')}\right]={\rm E}_{\bm{\omega}}\left[\langle e^{j\bm{\omega}^\intercal\mathbf{x}},e^{j\bm{\omega}^\intercal\mathbf{x}'}\rangle\right]\label{eq:Fourier_kernel}
\end{align}
where $\langle \cdot,\cdot \rangle$ is the Hermitian inner product $\langle \mathbf{x},\mathbf{x}'\rangle =\sum_i\mathbf{x}_i\overline{\mathbf{x}'_i}$, and $\langle e^{j\bm{\omega}^\intercal\mathbf{x}},e^{j\bm{\omega}^\intercal\mathbf{x}'}\rangle$ is an unbiased estimate of the properly scaled shift-invariant kernel $k(\mathbf{x}-\mathbf{x}')$ when $\bm{\omega}$ is drawn from the probability distribution $p(\bm{\omega})$. 

Typically, we ignore the imaginary part of the complex exponentials to obtain a real-valued mapping. This can be further generalized by considering the class of kernel functions with the following construction
\begin{align}
k(\mathbf{x},\mathbf{x}')=\int_{\mathcal{V}}z(\mathbf{v},\mathbf{x})z(\mathbf{v},\mathbf{x}')p(\mathbf{v})d\mathbf{v}
\end{align}
where $z:\mathcal{V}\times\mathcal{X}\rightarrow\R$ is a continuous and bounded function with respect to $\mathbf{v}$ and $\mathbf{x}$. The kernel function can be approximated by its Monte-Carlo estimate
\begin{align}
\tilde{k}(\mathbf{x},\mathbf{x}')=\frac{1}{D}\sum^{D}_{i=1}z(\mathbf{v}_i,\mathbf{x})z(\mathbf{v}_i,\mathbf{x}')
\end{align} 
with feature space $\mathbb{R}^D$ and $\{\mathbf{v}_i\}^D_{i=1}$ sampled independently from the spectral measure.

\subsection{Random Features}
In order to avoid the computation of a full $n\times n$ pairwise kernel matrix to solve learning problems on $n$ input samples, the continuous shift-invariant kernel can be approximated as $k(\mathbf{x},\mathbf{x}') : \X \times \X \to \R \approx z(\mathbf{x})^\intercal z(\mathbf{x}')$, where $z : \X \to \R^D$. Two embeddings are presented in \cite{rahimi2007RFF},
based on the Fourier transform $p(\omega)$ of the kernel $k$:
\[
z_{RFF1}(\mathbf{x}) := 
\sqrt{\frac{2}{D}}
\begin{bmatrix}
\sin(\bm{\omega}^\intercal_1 \mathbf{x})
\\ \cos(\bm{\omega}^\intercal_1 \mathbf{x})
\\ \vdots
\\ \sin(\bm{\omega}^\intercal_{D/2} \mathbf{x})
\\ \cos(\bm{\omega}^\intercal_{D/2} \mathbf{x})
\end{bmatrix}
,\;
\bm{\omega}_i\sim p(\bm{\omega})
\label{eq:z-tilde}
\]
and since both $p(\bm{\omega})$ and $k(\delta)$, where $\delta = \x-\x'$, are real, the integral in (\ref{eq:kintergral}) converges when cosines replace the complex exponentials
\[
z_{RFF2}(\mathbf{x}) := 
\sqrt{\frac{2}{D}}
\begin{bmatrix}
\cos(\bm{\omega}^\intercal_1 \mathbf{x} + b_1)
\\ \vdots
\\ \cos(\bm{\omega}^\intercal_D \mathbf{x} + b_D)
\end{bmatrix}
,\;
\begin{aligned}
\bm{\omega}_i &\sim p(\bm{\omega})
\\ b_i &\sim \rm{U}{[0, 2\pi]}
\end{aligned}
\label{eq:z-breve}
\]
where $\bm{\omega}$ is drawn from $p(\bm{\omega})$ and $b_i$ is drawn uniformly from $[0,2\pi]$. 

Let $\tilde{k}_{RFF1}$ and $\tilde{k}_{RFF2}$ be the reconstructions from $z_{RFF1}$ and $z_{RFF2}$ respectively and apply Ptolemy's trigonometric identities, we obtain
\begin{align}
\tilde{k}_{RFF1}(\x, \x')
=& \frac{1}{D/2} \sum_{i=1}^{D/2} \cos(\bm{\omega}_i^\intercal  (\mathbf{x} - \mathbf{x}'))
\label{eq:s-tilde}
\\    \tilde{k}_{RFF2}(\x, \x')
=& \frac{1}{D} \sum_{i=1}^D \bigl\{\cos(\bm{\omega}_i^\intercal (\mathbf{x} - \mathbf{x}'))\bigr.\nonumber\\
&\qquad\quad \bigl. + \cos(\bm{\omega}_i^\intercal (\mathbf{x} + \mathbf{x}') + 2b_i)\bigr\}.
\label{eq:s-breve}
\end{align}
Since ${\rm E}_{\bm{\omega}}{\rm E}_b\left[\cos(\bm{\omega}^\intercal (\mathbf{x} + \mathbf{x}') + 2b)\right] = 0$, both reconstructions are a mean of bounded terms with expected value $k(\mathbf{x}, \mathbf{x}')$. For a embedding dimension $D$, it is not immediately obvious which estimate is preferable: $z_{RFF2}$ results in twice as many samples for $\omega$, but introduces additional phase shift noise (non-shift-invariant) . In fact, all of the existing RFF kernel adaptive filtering in the literature, that we are aware of, uses $z_{RR2}$. However, it has been shown that this widely used variant has strictly higher variance for the Gaussian kernel with worse bounds \cite{Sutherland2015}:
\begin{align}
{\rm Var}[z_{RFF1}(\delta)] &= \frac{1}{D}[1+k(2\delta)-2k(\delta)^2]\\
{\rm Var}[z_{RFF2}(\delta)] &= \frac{1}{D}[1+\frac{1}{2}k(2\delta)-k(\delta)^2]
\end{align}
and $(1/2)k(2\delta)< k(\delta)^2$ for Gaussian kernel $k(\delta)=\exp(-\norm{\delta}^2/(2\sigma^2))$.
\subsection{Deterministic Features}
The RFF approach can be viewed as performing numerical integration using randomly selected sample points. We will explore ways to approximate the integral with a discrete sum of judiciously selected points.
\subsubsection{Gaussian Quadrature (GQ)}
There are many quadrature rules, without loss of generality, we focus on Gaussian quadrature, specifically the Gauss-Hermite quadrature using Hermite polynomials. Here, we present a brief summary (for a detailed discussion and proofs, please refer to \cite{Dao2017}). The objective is to select the abscissas $\bm{\omega}_i$ and corresponding weights $a_i$ to uniformly approximate the integral \eqref{eq:kintergral} with $\tilde{k}(\mathbf{x} - \mathbf{x}') =
\sum_{i=1}^D a_i \exp(j \bm{\omega}_i^\top(\mathbf{x} - \mathbf{x}'))$.
To obtain a \emph{feature map} $z: \X \rightarrow \C^D$ where $\tilde{k} (\mathbf{x} - \mathbf{x}') =
\sum_{i=1}^D a_i z_i(\mathbf{x}) \overline{z_i(\mathbf{x}')}$, we can define
\begin{equation*}
z(\mathbf{x}) = \begin{bmatrix} \sqrt{a_1} \exp(j \bm{\omega}_1^\top \mathbf{x}) & \dots & \sqrt{a_D} \exp(j
\bm{\omega}_D^\top \mathbf{x}) \end{bmatrix}^\top.
\end{equation*}
The maximum error for $\mathbf{x}, \mathbf{x}'$ in a region $\mathcal{M}$ with diameter $M = \sup_{\mathbf{x}, \mathbf{x}' \in \mathcal{M}} \norm{\mathbf{x}-\mathbf{x}'}$ is defined as
\begin{align}
\epsilon &= \sup_{(\mathbf{x}, \mathbf{x}') \in \mathcal{M}} \left |k(\mathbf{x}-\mathbf{x}') - \tilde{k}(\mathbf{x}-\mathbf{x}')\right |\nonumber\\
&= \sup_{\norm{\mathbf{x}-\mathbf{x}'} \le M} \left|\int_{\R^d} p(\bm{\omega}) e^{j \bm{\omega}^\intercal (\mathbf{x}-\mathbf{x}')} \, d \bm{\omega} - \sum_{i=1}^D a_i e^{j \bm{\omega}_i^\intercal (\mathbf{x}-\mathbf{x}')}\right|.
\label{eqnMaximumError}
\end{align}

A \emph{quadrature rule} is a choice of points $\omega_i$ and weights $a_i$ to minimize the maximum error $\epsilon$. For a fixed diameter $M$, the \emph{sample complexity} (SC) is defined as:
\begin{defn}
	\label{defnSampleComplexity}
	For any $\epsilon > 0$, a quadrature rule has sample complexity $D_{\mathrm{SC}}(\epsilon) = D$, where $D$ is the smallest number of samples such that the rule yields a maximum error of at most $\epsilon$.
\end{defn}
In numerical analysis, Gaussian quadrature is an exact-polynomial approximation of a one-dimensional definite integral:  $\int p(\bm{\omega}) f(\bm{\omega}) \, d \bm{\omega} \approx \sum_{i=1}^D a_i f(\bm{\omega}_i)$, where the $D$-point construction yields an exact result for polynomials of degree up to $2D - 1$. While the GQ points and corresponding weights are both distribution $p(\bm{\omega})$ and parameter $D$ dependent, they can be computed efficiently using orthogonal polynomials. GQ approximations are accurate for integrating functions that are well-approximated by polynomials, including all sub-Gaussian densities.
\begin{defn}[Sub-Gaussian Distribution]
	A distribution $p: \R^d \rightarrow \R$ is b-\emph{sub-Gaussian}, with parameter $b$, if for $X \sim p$ and for all $t \in \R^d$, ${\rm E}[\exp(\langle t, X \rangle)] \le \exp\left(
	\frac{1}{2} b^2 \norm{t}^2 \right)$.
\end{defn}
This is more restrictive compared to the RFFs, but we are primarily concerned with the Gaussian kernel in this paper, and any kernel can be approximated by its convolution with the Gaussian kernel, resulting in a much smaller approximation error than the data generation noise \cite{Dao2017}.

To extend one-dimensional GQ (accurate for polynomials up to degree $R$) to higher dimensions, the following  must be satisfied
\begin{align}
\int_{\R^d} p(\bm{\omega}) \prod_{l=1}^d 
(\mathbf{e}_l^\intercal\bm{\omega})^{r_l} d\bm{\omega} = \sum_{i=1}^D a_i \prod_{l=1}^d (\mathbf{e}_l^\intercal \bm{\omega}_i)^{r_l}\label{eqnPolyExactConstraints}
\end{align}
for all $r \in \mathbb{N}^d$ such that $\sum_l r_l \leq R$, where $\mathbf{e}_l$ are the standard basis vectors.

For sub-Gaussian kernels, the maximum error of an exact-polynomial feature map can be bounded using the Taylor series approximation of the
exponential function in \eqref{eqnMaximumError}.
\begin{thm}
	\label{thmPolynomialQuadrature}
	Let $k$ be a $b$-sub-Gaussian kernel, $\tilde{k}$ be its
	quadrature rule estimate with non-negative weights that is exact up
	to some even degree $R$, and $\mathcal{M} \subset \R^d$ be some region with diameter $M$, then, for all $\mathbf{x}, \mathbf{x}' \in \mathcal{M}$, the approximation error is bounded by
	\begin{equation*}
	\abs{k(\mathbf{x}- \mathbf{x}') - \tilde{k}(\mathbf{x} - \mathbf{x}')} \le 3 \left(\frac{e b^2 M^2}{R}
	\right)^{\frac{R}{2}}.
	\end{equation*}
\end{thm}
There are $\binom{d + R}{d}$ constraints in \eqref{eqnPolyExactConstraints}, thus to satisfy the conditions of Theorem 2, we need to use at least this many sample points.

\begin{corollary}
	\label{corPolynomialQuadrature}
	For a class of feature maps satisfying the conditions in
	Theorem~\ref{thmPolynomialQuadrature}, and have sample size $D
	\le \beta {d + R \choose d}$ for some fixed constant $\beta$.
	Then, for any $\gamma > 0$, the sample complexity of features maps in this class is bounded by
	\begin{equation*}
	D(\epsilon)
	\le \beta 2^d \max\left(\exp\left(e^{2 \gamma + 1} b^2 M^2 \right), \left( \frac{3}{\epsilon}
	\right)^{\frac{1}{\gamma}} \right).
	\end{equation*}
	In particular, for a fixed dimension $d$, this means that for any $\gamma$, sample complexity $D_{SC}(\epsilon) =
	O\left(\epsilon^{-\frac{1}{\gamma}} \right)$.
\end{corollary}

Compared to random Fourier features, with $D_{SC}(\epsilon) = O(\epsilon^{-2})$, GQ features have a much weaker dependence on $\epsilon$, at a constant cost of an additional factor of $2^d$ (independent of the error $\epsilon$).

To construct GQ features, we can randomly select points $\bm{\omega}_i$, then solve \eqref{eqnPolyExactConstraints} for the weights $a_i$ using non-negative least squares (NNLS) algorithm. This works well in low dimensions, but for higher values of d and R, grid-based quadrature rules can be constructed efficiently.

A \emph{dense grid} or tensor-product construction factors the integral \eqref{eq:kintergral} along the dimensions
$k(u) = \prod_{i=1}^d \left(\int_{-\infty}^{\infty} p_i(\omega)
\exp(j \omega e_i^\intercal u) \, d \omega \right)$, where $e_i$ are the standard basis vectors, and can be approximated using one-dimensional quadrature rule. However, since the sample complexity is doubly-exponential in $d$, a \emph{sparse grid} or Smolyak quadrature is used \cite{Smolyak63}. Only points up to some fixed total level A is included, achieving a similar error with exponentially fewer points than a single larger quadrature rule. 

The major drawback of the grid-based construction is the lack of fine tuning for the feature dimension. Since the number of samples extracted in the feature map is determined by the degree of polynomial exactness, even a small incremental change can produce a significant increase in the number of features. \emph{Subsampling} according to the distribution determined by their weights is used to combat both the curse of dimensionality and the lack of detailed control over the exact sample number. There are also data-adaptive methods to choose a quadrature rule for a predefined number of samples \cite{Dao2017}.

\subsubsection{Taylor Polynomials (TS)}\label{sec:taylor_features}
Here, we present a brief summary of the explicit Taylor series (TS) expansion polynomial features for the Gaussian kernel with respect to the inner product $\langle \mathbf{x}, \mathbf{x}'\rangle$ (for a detailed discussion, please refer to \cite{cotter2011explicit}). This is the most straightforward deterministic feature map for the Gaussian kernel, where each term in the TS is expressed as a sum of matching monomials in $\mathbf{x}$ and $\mathbf{x}'$, i.e.,
\begin{equation}\label{eq:express_Gaussian}
k(\mathbf{x}, \mathbf{x}') = e^{-\frac{\|\mathbf{x}- \mathbf{x}'\|^2}{2\sigma^2}} =
e^{-\frac{\|\mathbf{x}\|^2}{2\sigma^2}}e^{-\frac{\|\mathbf{x}'\|^2}{2\sigma^2}}e^{\frac{\langle \mathbf{x}, \mathbf{x}'\rangle}{\sigma^2}}.
\end{equation}
We can easily factor out the first two product terms that depend on $\x$ and $\x'$ independently. The last term in \eqref{eq:express_Gaussian}, $e^{\frac{\langle \mathbf{x}, \mathbf{x}'\rangle}{\sigma^2}}$, can be expressed as a power series or infinite sum using Taylor polynomials as
\begin{equation}\label{eq:scalar_taylor}
e^{\frac{\inner{\x,\x'}}{\sigma^2}} = \sum_{n=0}^{\infty}
\frac{1}{n!}\left(\frac{\inner{\x,\x'}}{\sigma^2}\right)^n.
\end{equation}
Using shorthand, we can factor the inner-product exponentiation as
\begin{equation}\label{eq:multinomial_exp}
\inner{\x,\x'}^n=\left(\sum_{i=1}^d \x_i \x'_i\right)^n=\sum_{j \in [d]^n} \left(
\prod_{i=1}^n \x_{j} \right) \left( \prod_{i=1}^n \x'_{j} \right)
\end{equation}
where $j$ enumerates over all selections of $d$ coordinates (including repetitions and different orderings of the same coordinates) thus avoiding collecting equivalent terms and writing down their corresponding multinomial coefficients, i.e., as an inner product between degree $n$ monomials of the coordinates of $\x$ and $\x'$. Substituting this into \eqref{eq:scalar_taylor} and \eqref{eq:express_Gaussian} yields the following explicit feature map:
\begin{equation}
\label{eq:taylor_features} \hat{z}_{n,j}\left( \x \right) =
e^{-\frac{\|\x\|^2}{2\sigma^2}} \frac{1}{\sigma^{n}\sqrt{n!}}\prod_{i=0}^{n}
{\x}_{j}
\end{equation}
where $k(\x,\x') = \inner{\hat{z}(\x),\hat{z}(\x')} = \prod_{k=0}^{\infty} \prod_{j\in [d]^k} \hat{z}_{n,j}(\x)\hat{z}_{n,j}(\x')$. For TS feature approximation, we truncate the infinite sum to the first $r+1$ terms:
\begin{equation}
\label{eq:taylor_kernel_expansion} \tilde{k}(\x,\x')=\inner{\hat{z}(\x),\hat{z}(\x')}
= e^{-\frac{\|\x\|^2+\|\x'\|^2}{2\sigma^2}} \sum_{n=0}^{r} \frac{1}{n!}\left(\frac{\inner{\x,\x'}}{\sigma^2}\right)^n
\end{equation}
where the TS approximation is exact up to polynomials of degree $r$.
\begin{thm}
	The error of the Taylor series features approximation is bounded by	
	\begin{equation*}
	\abs{k(\x,\x')-\tilde{k}(\x,\x')}\le \frac{1}{(r+1)!} \left(
	\frac{\|\x\|\,\|\x'\|}{\sigma^2} \right)^{r+1}.
	\end{equation*}
\end{thm}
In practice, the different permutations of j in each $n$-th term of Taylor expansion, \eqref{eq:multinomial_exp}, are collected into a single feature corresponding to a distinct monomial, resulting in ${{d+n-1} \choose {n}}$ features of degree $n$, and a total of $D={{d+r} \choose {r}}$ features of degree at most $r$. Similar to GQ, the number of samples extracted in the feature map is determined by the number of terms in the expansion, where a small change in the input dimension $d$ can yield a significant increase in the number of features. To overcome this complexity, we can employ a similar sub-sampling scheme as GQ. Polynomials are prone to numerical instability. It is important to properly scale the data first, before applying a polynomial kernel. The range of $\pm1$ has been shown to be effective for our simulation. A square-root formulation to reduce numerical error by improving precision and stability can also be used. 
\subsection{Polynomial vs. Fourier Decomposition}
Here, we make some observations regarding the random and deterministic feature constructions presented in this paper.

The Fourier series (generalized by the Fourier transform) express a function as an infinite sum of exponential functions (sines and cosines). Taylor series and Gaussian quadrature belong to a class of polynomial decomposition that expresses a function as an infinite sum of powers.

The finite truncations of the Taylor series of a function defined by a power series of the form $f(x)=\sum^\infty_{n=0}c_n(x-a)^n$, about the point $x = a$, are all exactly equal to $f(a)$. In contrast, the Fourier coefficients are computed by integrating over an entire interval, so there is generally no such point where all the finite truncations of the series are exact.

The computation of Taylor polynomials requires the knowledge of the function on an arbitrary small neighborhood of a point (local), whereas the Fourier series requires knowledge on the entire domain (global). Thus, for the Taylor series, the error is very small in a neighborhood of the point that is computed (while it may be very large at a distant point). On the other hand, for Fourier series, the error is distributed along the domain of the function.

In fact, the Taylor and Fourier series are equivalent in the complex domain, i.e., if we restrict the complex variable to the real axis, we obtain the Taylor series, and if we restrict the complex variable to the unit circle, we obtain the Fourier series. Thus, real-valued Taylor and Fourier series are specific cases of complex Taylor series. As discussed above, a duality exists: local properties (derivatives) vs. global properties (integrals over circles). The coefficients of TS are computed by differentiating a function $n$ times at the point $x=0$, while the Fourier series coefficients are computed by integrating the function multiplied by a sinusoidal wave, oscillating $n$ times.

\subsection{Universal Approximation}
Finite dimensional feature space mappings such as random Fourier features, Gaussian quadrature, and Taylor polynomials are often viewed as an approximation to a kernel function $k(\x-\x')=\langle \phi(\x),\phi(\x')\rangle_{\mathcal{H}}$. However, it is more appropriate to view them as an equivalent kernel that defines a new reproducing kernel Hilbert space: a nonlinear mapping $z(\cdot)$ that transforms the data from the original input space to a new higher finite-dimensional RKHS $\mathcal{H}'$ where $k'(\x-\x')=\langle z(\x),z(\x')\rangle_{\mathcal{H}'}$. The RKHS $\mathcal{H}'$ is not necessarily contained in the RKHS $\mathcal{H}$ corresponding to the kernel function $k$, e.g., Gaussian kernel. It is easy to show that the mappings discussed in this paper induce a positive definite kernel function satisfying Mercer's conditions. 
\begin{pro}[Closer properties]
	Let $k_1$ and $k_2$ be positive-definite kernels over $\mathcal{X}\times\mathcal{X}$ (where $\mathcal{X}\subseteq\R^d$), $a\in \R^+$ is a positive real number, $f(\cdot)$ a real-valued function on $\mathcal{X}$, then the following functions are positive definite kernels.
	\begin{enumerate}
		\item $k(\x,\y) = k_1(\x,\y)+k_2(\x,\y)$,
		\item $k(\x,\y) = ak_1(\x,\y)$,
		\item $k(\x,\y) = k_1(\x,\y)k_2(\x,\y)$,
		\item $k(\x,\y) = f(\x)f(\y)$.
	\end{enumerate}
\end{pro}
Since exponentials and polynomials are positive-definite kernels, under the closer properties, it is clear that the dot products of random Fourier features, Gaussian quadrature, and Taylor polynomials are all reproducing kernels. It follows that these kernels have universal approximating property: approximates uniformly an arbitrary continuous target function to any degree of accuracy over any compact subset of the input space.

\section{Kernel Adaptive Filtering (KAF)}\label{Sec:KAF}
Kernel methods \cite{Scholkopf01}, including support vector machine (SVM) \cite{Vapnik95}, kernel principal component analysis (KPCA) \cite{Scholkopf98}, and Gaussian process (GP) \cite{Rasmussen06}, form a powerful unifying framework for classification, clustering, PCA, and regression, with many important applications in science and engineering. In particular, the theory of adaptive signal processing is greatly enhanced through the integration of the theory of reproducing kernel Hilbert space. By performing classical linear methods in a potentially infinite-dimensional feature space, KAF \cite{Liu10} moves beyond the limitations of the linear model to provide general nonlinear solutions in the original input space. This bridges the gap between adaptive signal processing and artificial neural networks (ANNs), combining the universal approximation property of neural networks and the simple convex optimization of linear adaptive filters. Since its inception, there have been many important recent advancements in KAF, including the first auto-regressive model built on a full state-space representation \cite{KAARMA} and the first functional Bayesian filter \cite{li2019functional}. 
\subsection{Kernel Least Mean Square (KLMS)}
First, we briefly discuss the simplest of KAF algorithms, the kernel least mean square (KLMS) \cite{KLMS}. In machine learning, supervised learning can be grouped into two broad categories: classification and regression. For a set of $N$ data points $\mathcal{D} = \{\x_n,y_n\}^{N}_{n=1}$, the desired output $y$ is either categorical variables, e.g., $y\in\{-1,+1\}$, in the case of binary classification, or real numbers, e.g., $y\in\mathbb{R}$, for the task of regression or interpolation, where $\textbf{X}^{N}_1\stackrel{\Delta}{=} \{\x_n\}^{N}_{n=1}$ is the set of $d$-dimensional input vectors, i.e., $\x_n\in\mathbb{R}^d$, and $\textbf{y}^{N}_1 \stackrel{\Delta}{=} \{y_n\}^{N}_{n=1}$ is the corresponding set of desired signal or observations. In this paper, we will focus on the latter problem, although the same approach can be easily extended for classification. The task is to infer the underlying function $y=f(\x)$ from the given data  $\mathcal{D} = \{\textbf{X}^{N}_1,\textbf{y}^{N}_1\}$ and predict its value, or the value of a new observation $y'$, for a new input vector $\x'$. Note that the desired data may be noisy in nature, i.e., $y_n = f(\x_n) + v_n$, where $v_n$ is the noise at time $i$, which we assume to be independent and identically distributed (i.i.d.) Gaussian random variable with zero-mean and unit-variance, i.e., $V\sim\mathcal{N}(0,1)$.

For a parametric approach or weight-space view to regression, the estimated latent function $\hat{f}(\x)$ is expressed in terms of a parameters or weights vector $\textbf{w}$. In the standard linear form
\begin{equation}
\hat{f}(\x)=\textbf{w}^\intercal\x.\label{eq:paramodel}
\end{equation} 
To overcome the limited expressiveness of this model, we can project the $d$-dimensional input vector $\x\in\mathbb{U}\subseteq\mathbb{R}^d$ (where $\mathbb{U}$ is a compact input domain in $\mathbb{R}^d$) into a potentially infinite-dimensional feature space $\mathbb{F}$. Define a $\mathbb{U}\rightarrow\mathbb{F}$ feature mapping $\phi(\x)$, the parametric model (\ref{eq:paramodel}) becomes
\begin{equation}
\hat{f}(\x) = \bm{\Omega}^\intercal\phi(\x)\label{FSR}
\end{equation} 
where $\bm{\Omega}$ is the weight vector in the feature space. 

Using the Representer Theorem \cite{RT} and the \emph{kernel trick}, (\ref{FSR}) can be expressed as
\begin{equation}
\hat{f}(\x)=\sum^N_{n=1}\alpha_n k(\x_n,\x)\label{eq:KernelTrick}
\end{equation}
where $k(\x,\x')$ is a Mercer kernel, corresponding to the inner product $\left\langle\phi(\x), \phi(\x')\right\rangle$, and $N$ is the number of basis functions or training samples. Note that $\mathbb{F}$ is equivalent to the reproducing kernel Hilbert spaces (RKHS) induced by the kernel if we identify $\phi(\x)= k(\x,\cdot)$. The most commonly used kernel is the Gaussian or radial basis kernel
\begin{equation}
k_\sigma(\x,\x') = \exp\left(-\frac{\norm{\x-\x'}^2}{2\sigma^2}\right)\label{eq:GaussKernel}
\end{equation} 
where $\sigma>0$ is the kernel parameter, and can be equivalently expressed as $k_a(\x,\x') = \exp\left(-a\norm{\x-\x'}^2\right)$, where $a = 1/(2\sigma^2)$.

The learning rule for the KLMS algorithm in the feature space follows the classical linear adaptive filtering algorithm, the least mean square
\begin{align}
\left\{ 
\begin{array}{l}
\boldsymbol\Omega_0 = \textbf{0}\\
e_n = y_n - \left\langle\boldsymbol\Omega_{n-1}, \phi(\x_n)\right\rangle\\
\boldsymbol\Omega_n = \boldsymbol\Omega_{n-1} + \eta e_n \phi(\x_n)
\end{array} \right.
\end{align}
which, in the original input space, becomes
\begin{align}
\left\{ 
\begin{array}{l}
\hat{f}_0 = 0\\
e_n = y_n - \hat{f}_{n-1}(\x_n)\\
\hat{f}_n = \hat{f}_{n-1} + \eta e_n k(\x_n,\cdot)\\
\end{array} \right.
\end{align}
where $e_n$ is the prediction error in the $n$-th time step, $\eta$ is the learning rate or step-size, and $f_n$ denotes the learned mapping at iteration $n$. Using KLMS, we can estimate the mean of $y$ with linear per-iteration computational complexity $O(n)$, making it an attractive online algorithm. 
\subsection{Kernel Recursive Least Square (KRLS)}
The LMS and similar adaptive algorithms using stochastic gradient descent aim to reduce the mean squared error (MSE) based on the current error. Recursive least square (RLS), on the other hand, recursively finds the filter parameters that minimize a weighted linear least squares cost function with respect to the input signals, based on the total error. Hence, the RLS exhibits extremely fast convergence, however, at the cost of high computational complexity. We first present the linear RLS, followed by its kernelized version.

The exponentially weighted RLS filter minimizes a cost function $C$ through the filter coefficients $\mathbf{w}_n$ in each iteration
\begin{equation}
C(\mathbf{w}_n)=\sum_{i=1}^{n}\lambda^{n-i}e^{2}_i
\end{equation}
where the prediction error is $e_{n}=y_{n}-\hat{y}_{n}$ and $0<\lambda\le 1$ is the ``forgetting factor'' which gives exponentially less weight to earlier error samples.

The cost function is minimized by taking the partial derivatives for all entries $k$ of the weight vector $\mathbf{w}_{n}$ and setting them to zero
\begin{equation}
\frac{\partial C(\mathbf{w}_{n})}{\partial \w_{n}}=\sum_{i=1}^{n}\,2\lambda^{n-i}e_i\,\frac{\partial e_i}{\partial \w_{n}}={-}\sum_{i=1}^{n}\,2\lambda^{n-i}e_i\,\x^\intercal_{i}=0. 
\end{equation}

Substituting $e_{i}$ for its definition yields
\begin{equation}
\sum_{i=1}^{n}\lambda^{n-i}\left(y_i-\w^\intercal_{n}\x_{i}\right)\x^\intercal_{i}= 0.
\end{equation}
Rearranging the terms, we obtain
\begin{equation}
\w^\intercal_{n}\left[\sum_{i=1}^{n}\lambda^{n-i}\x_{i}\x^\intercal_{i}\right]= \sum_{i=1}^{n}\lambda^{n-i}y_i\x^\intercal_{i}.
\end{equation}
This can be expressed as $\mathbf{R}_{x}(n)\,\mathbf{w}_{n}=\mathbf{r}_{yx}(n)$,
where $\mathbf{R}_{x}(n)$ and $\mathbf{r}_{yx}(n)$ are the weighted sample covariance matrix for $x_{n}$, and the cross-covariance between $d_{n}$ and $x_{n}$, respectively. Thus, the cost-function minimizing coefficients are
\begin{equation}
\mathbf{w}_{n}=\mathbf{R}_{x}^{-1}(n)\,\mathbf{r}_{yx}(n).\label{eq:RLSweight}
\end{equation}
The covariance and cross-covariance can be recursively computed as
\begin{equation}
\mathbf{R}_{x}(n)=\sum_{i=1}^{n}\lambda^{n-i}\mathbf{x}_i\mathbf{x}^\intercal_i=\lambda\mathbf{R}_{x}(n-1)+\mathbf{x}_{n}\mathbf{x}^\intercal_{n}\label{eq:RLScovariance}
\end{equation}
and
\begin{equation}
\mathbf{r}_{yx}(n)=\sum_{i=1}^{n}\lambda^{n-i}y_i\mathbf{x}_i=\lambda\mathbf{r}_{dx}(n-1)+y_{n}\mathbf{x}_{n}.\label{eq:RLScrosscovariance}
\end{equation}
Using the shorthand $\mathbf{P}_{n}=\mathbf{R}_{x}^{-1}(n)$ and applying the matrix inversion lemma (specifically, the Sherman–Morrison formula, with $\mathbf{A} =\lambda\mathbf{P}^{-1}_{n-1}$, $\mathbf{u}=\mathbf{x}_{n}$, and $\mathbf{v}^\intercal=\mathbf{x}^{T}_{n}$)  to \eqref{eq:RLScovariance} yields
\begin{align}
\mathbf{P}_{n}=&\left[\lambda\mathbf{P}^{-1}_{n-1}+\mathbf{x}_{n}\mathbf{x}^{T}_{n}\right]^{-1}\nonumber\\
=&\lambda^{-1}\mathbf{P}_{n-1}
-\frac{\lambda^{-1}\mathbf{P}_{n-1}\mathbf{x}_{n}\mathbf{x}^{T}_{n}\lambda^{-1}\mathbf{P}_{n-1}}{1+\mathbf{x}^{T}_{n}\lambda^{-1}\mathbf{P}_{n-1}\mathbf{x}_{n}}.\label{eq:RLSinversion}
\end{align}
From \eqref{eq:RLSweight} and \eqref{eq:RLScrosscovariance}, the weight update recursion becomes
\begin{equation}
\mathbf{w}_{n}=\mathbf{P}_{n}\,\mathbf{r}_{dx}(n)=\lambda\mathbf{P}_{n}\,\mathbf{r}_{dx}(n-1)+y_{n}\mathbf{P}_{n}\,\mathbf{x}_{n}.
\end{equation}
Substituting \eqref{eq:RLSinversion} into the above expression and through simple manipulations, we obtain
\begin{equation}
\mathbf{w}_{n}=\mathbf{w}_{n-1}+\mathbf{g}_{n}\left[d_{n}-\mathbf{w}^{T}_{n-1}\mathbf{x}_{n}\right]=\mathbf{w}_{n-1}+\mathbf{g}_{n}e^{(-)}_n
\end{equation}
where $\mathbf{g}_{n}=\mathbf{P}_{n}\mathbf{x}_{n}$ is the gain and $e^{(-)}_n=y_{n}-\mathbf{w}^{T}_{n-1}\mathbf{x}_{n}$ is the \emph{a priori} error (the \emph{a posteriori} error is  $e^{(+)}_{n}=y_{n}-\mathbf{w}^{T}_n\mathbf{x}_{n}$). The weights correction factor is
$\Delta\mathbf{w}_{n-1}=\mathbf{g}_{n}e^{(-)}_n$, which is directly proportional to both the prediction error and the gain vector, which controls the desired amount of sensitivity is desired, through the forgetting factor, $\lambda$.

The RLS is equivalent to the Kalman filter estimate of the weight vector for the following system
\begin{align}
\w_{n+1} &= \w_{n} &\makebox{\qquad(state model)}\\
d_{n} &=\w_{n}^\intercal\x_{n}+v_{n} &\makebox{\qquad(observation model)}
\end{align}
where the observation noise is assumed to be additive white Gaussian, i.e., $v_n\sim\mathcal{N}(0,\sigma_v^2)$. For stationary and slow-varying nonstationary processes, the RLS convergences extremely fast.  We can improve the tracking ability of the RLS algorithm with the addition of a state transition matrix
\begin{align}
\w_{n+1} &= \mathbf{A}\w_{n} + \mathbf{q}_{n} &\makebox{\qquad(state model)}\\
d_{n} &=\w_{n}^\intercal\x_{n}+v_{n} &\makebox{\qquad(observation model)}
\end{align}
where the transition is typically set as $\mathbf{A} = \alpha \mathbf{I}$ for simplicity with zero-mean additive white Gaussian noise, i.e., process noise $\mathbf{q}_{n}\sim\mathcal{N}(0,q\mathbf{I})$. The kernel Ex-RLS or Ex-KRLS (Ex-RLS) resorts to two special cases of the KRLS (RLS): the random walk model ($\alpha = 1$) and the exponentially weighted KRLS ($\alpha = 1$ and $q = 0$).

The weight vector $\bm{\Omega_{n}}$ in kernel recursive least squares (KRLS) is derived using the following formulation
\begin{equation}
\bm{\Omega}_{n} = \left[\lambda \textbf{I}+\bm{\Phi}_{n}\bm{\Phi}_{n}^\intercal\right]^{-1}\bm{\Phi}_{n}\mathbf{y}_{n}\label{eq:omega_update}
\end{equation}
where $\mathbf{y}_{n}=[y_1,\cdots,y_{n}]^\intercal$ and $\bm{\Phi}_{n}=[\phi(\x_1),\cdots,\phi(\x_n)]$. Using the push-through identity, \eqref{eq:omega_update} can be express as
\begin{equation}
\bm{\Omega}_{n} = \bm{\Phi}_{n}\left[\lambda \textbf{I}+\bm{\Phi}_{n}^\intercal\bm{\Phi}_{n}\right]^{-1}\mathbf{y}_{n}.\label{eq:omega_update}
\end{equation}
Denote the inverse of the regularized Gram matrix as
\begin{equation}
\textbf{Q}_{n}=\left[\lambda \textbf{I}+\bm{\Phi}_{n}^\intercal\bm{\Phi}_{n}\right]^{-1}
\end{equation}
yielding
\begin{equation}
\textbf{Q}_{n}^{-1}=\begin{bmatrix}
\textbf{Q}_{n}^{-1} & \textbf{h}_{n}\\
\textbf{h}_{n}^\intercal & \lambda + \bm{\Phi}_{n}^\intercal\bm{\Phi}_{n}.
\end{bmatrix}
\end{equation}
where $\textbf{h}_{n}=\bm{\Phi}_{n-1}^\intercal\phi_{n}$. We obtain the following recursive matrix update
\begin{equation}
\textbf{Q}_{n}^{-1}=r_{n}^{-1}\begin{bmatrix}
\textbf{Q}_{n-1}r_{n} + \bm{\zeta}_{n}\bm{\zeta}_{n}^\intercal & -\bm{\zeta}_{n}\\
-\bm{\zeta}_{n} ^\intercal & 1
\end{bmatrix}
\end{equation}
where $\bm{\zeta}_{n} = \textbf{Q}_{n-1}\textbf{h}_{n}$ and $r_{n} =\lambda +\phi_{n}^\intercal\phi_{n}--\bm{\zeta}_{n} ^\intercal\textbf{h}_{n}$.
The KRLS is computationally expensive with space and time complexity $O(n^2)$. 
\subsection{No-Trick Kernel Adaptive Filtering (NT-KAF)}
Without loss of generality, we focus on the three most popular adaptive filtering algorithms discussed above. The same principle can be easily applied to any linear filtering techniques to achieve nonlinear performance in a higher finite-dimensional feature space using explicit mapping.

Once explicitly mapped into a higher finite-dimensional feature space where the dot product is a reproducing kernel, the linear LMS and RLS can be applied directly, with constant complexity $O(D)$ and $O(D^2)$, respectively. The NT-KLMS is summarized in Alg. \ref{alg:FS-LMS}, the NT-KRLS in Alg. \ref{alg:FS-RLS}, and the NT-Ex-KRLS in Alg. \ref{alg:FS-ExRLS}. 
\begin{algorithm}
	\textbf{Initialization:}\\
	$z(\cdot):\X\rightarrow\R^D$: feature map\\
	$\mathbf{w}(0) = \textbf{0}$: feature space weight vector $\mathbf{w}\in\R^D$\\
	$\eta$: learning rate\\
	\textbf{Computation:}\\
	\For{n = 1, 2, $\cdots$}{
		$e^{(-)}_n = y_{n} -\textbf{w}^\intercal_{n-1}z(\x_{n})$\\
		$\mathbf{w}_{n} = \mathbf{w}_{n-1}+\,\eta e^{(-)}_n$
	}
	\normalsize
	\caption{NT-KLMS Algorithm}
	\label{alg:FS-LMS}	
\end{algorithm}

\begin{algorithm}
	\textbf{Initialization:}\\
	$z(\cdot):\X\rightarrow\R^D$: feature map\\
	$\lambda$: forgetting factor\\
	$\delta$: initial value to seed the inverse covariance matrix $\mathbf{P}$\\
	$\mathbf{P}(0)=\delta\mathbf{I}$: where $\mathbf{I}$ is the $D\times D$ identity matrix\\
	$\mathbf{w}(0) = \textbf{0}$: feature space weight vector $\mathbf{w}\in\R^D$\\
	\textbf{Computation:}\\
	\For{n = 1, 2, $\cdots$}{
		$e^{(-)}_n = y_{n} -\textbf{w}^\intercal_{n-1}z(\x_{n})$\\
		$\mathbf{g}_{n}=\mathbf{P}_{n-1}z(\x_{n})\left\{\lambda+z(\x_{n})^\intercal\mathbf{P}_{n-1}z(\x_{n})\right\}^{-1}$\\	
		$\mathbf{P}_{n}=\lambda^{-1}\mathbf{P}_{n-1}-\mathbf{g}_{n}z(\x_{n})^\intercal\lambda^{-1}\mathbf{P}_{n-1}$\\
		$\mathbf{w}_{n} = \mathbf{w}_{n-1}+\,\mathbf{g}_{n}e^{(-)}_n$
	}
	\normalsize
	\caption{NT-KRLS Algorithm}
	\label{alg:FS-RLS}	
\end{algorithm}

\begin{algorithm}
	\textbf{Initialization:}\\
	$z(\cdot):\X\rightarrow\R^D$: feature map\\
	$\lambda$: forgetting factor\\
	$\mathbf{A}\in\R^{D\times D}$: state transition matrix\\
	$\delta$: initial value to seed the inverse covariance matrix $\mathbf{P}$\\
	$\mathbf{P}(0)=\delta\mathbf{I}$: where $\mathbf{I}$ is the $D\times D$ identity matrix\\
	$\mathbf{w}(0) = \textbf{0}$: feature space weight vector $\mathbf{w}\in\R^D$\\
	\textbf{Computation:}\\
	\For{n = 1, 2, $\cdots$}{
		$e^{(-)}_n = y_{n} -\textbf{w}^\intercal_{n-1}z(\x_{n})$\\
		$\mathbf{g}_{n}=\mathbf{A}\mathbf{P}_{n-1}z(\x_{n})\left\{\lambda+z(\x_{n})^\intercal\mathbf{P}_{n-1}z(\x_{n})\right\}^{-1}$\\	
		$\mathbf{P}_{n}=\mathbf{A}\left\{\lambda^{-1}\mathbf{P}_{n-1}-\mathbf{g}_{n}z(\x_{n})^\intercal\lambda^{-1}\mathbf{P}_{n-1}\right\}\mathbf{A}^\intercal+\lambda q\mathbf{I}{}$\\
		$\mathbf{w}_{n} = \mathbf{A}\mathbf{w}_{n-1}+\,\mathbf{g}_{n}e^{(-)}_n$
	}
	\normalsize
	\caption{NT-Ex-KRLS Algorithm}
	\label{alg:FS-ExRLS}
\end{algorithm}

Since points in the explicitly defined RKHS is related to the original input through a change of variables, all of the theorems, identities, etc. that apply to linear filtering have an NT-RKHS counterpart.
\section{Simulation Results}\label{Sec:Simulation}
	Here we perform one-step ahead prediction on the Mackey-Glass (MG) chaotic time series \cite{Mackey77}, defined by the following time-delay ordinary differential equation
	\begin{equation*}
	\frac{d x(t)}{d t} =\frac{\beta x(t-\tau)}{1+x(t-\tau)^{n}} -\gamma x(t)
	\end{equation*}
	where $\beta=0.2$, $\gamma=0.1$, $\tau=30$, $n=10$, discretized at a sampling period of 6 seconds using the forth-order Runge-Kutta method, with initial condition $x(t) = 0.9$. Chaotic dynamics are extremely sensitive to initial conditions: small differences in initial conditions yields widely diverging outcomes, rendering long-term prediction intractable, in general. 
\subsection{Chaotic Time Series Prediction using (K)LMS Algorithms}
\begin{figure}[t!]
	\centering
	\begin{subfigure}
		\centering
		\includegraphics[width=0.40\textwidth]{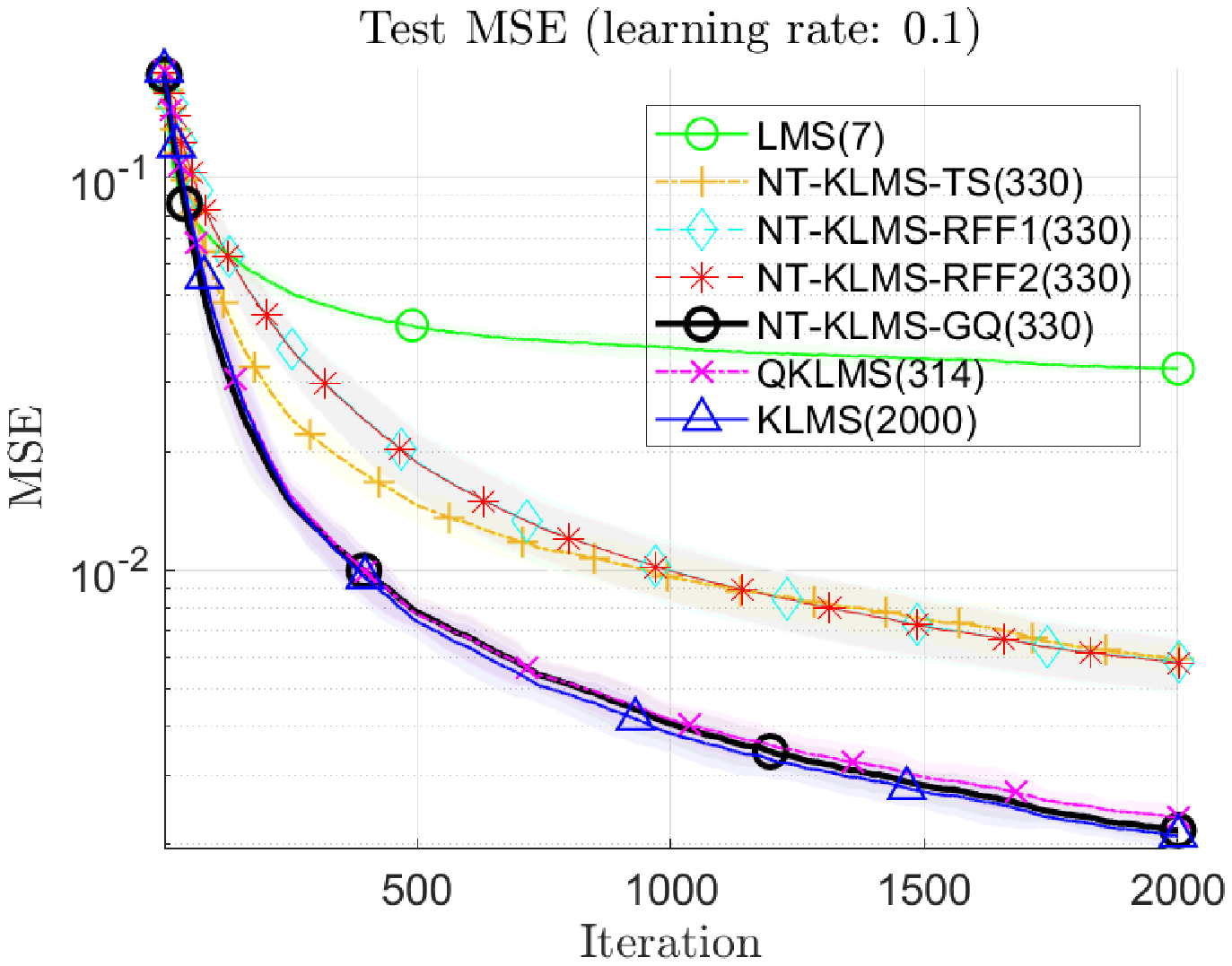}
	\end{subfigure}%
	\begin{subfigure}
		\centering
		\includegraphics[width=0.40\textwidth]{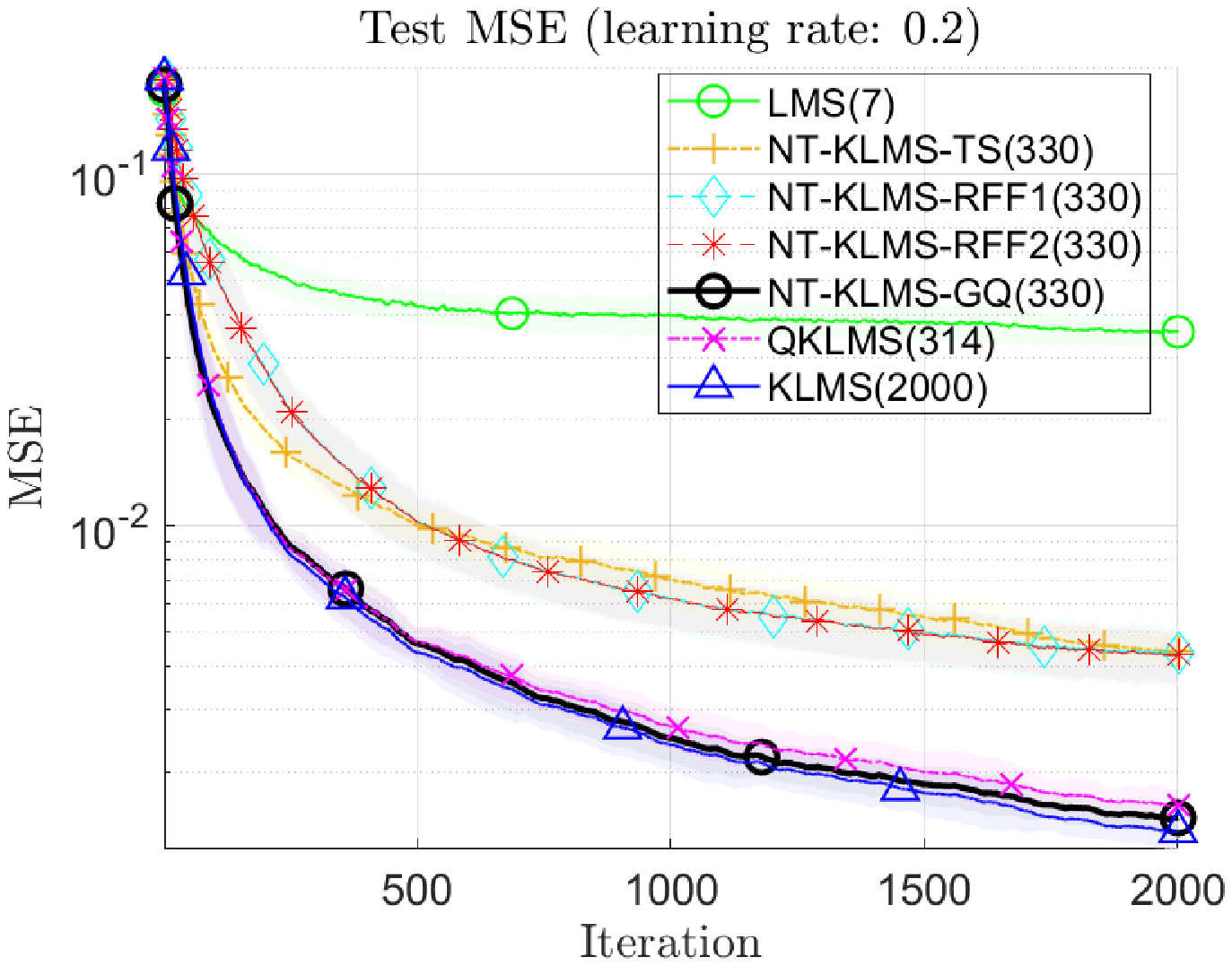}
	\end{subfigure}
	\begin{subfigure}
	\centering
	\includegraphics[width=0.40\textwidth]{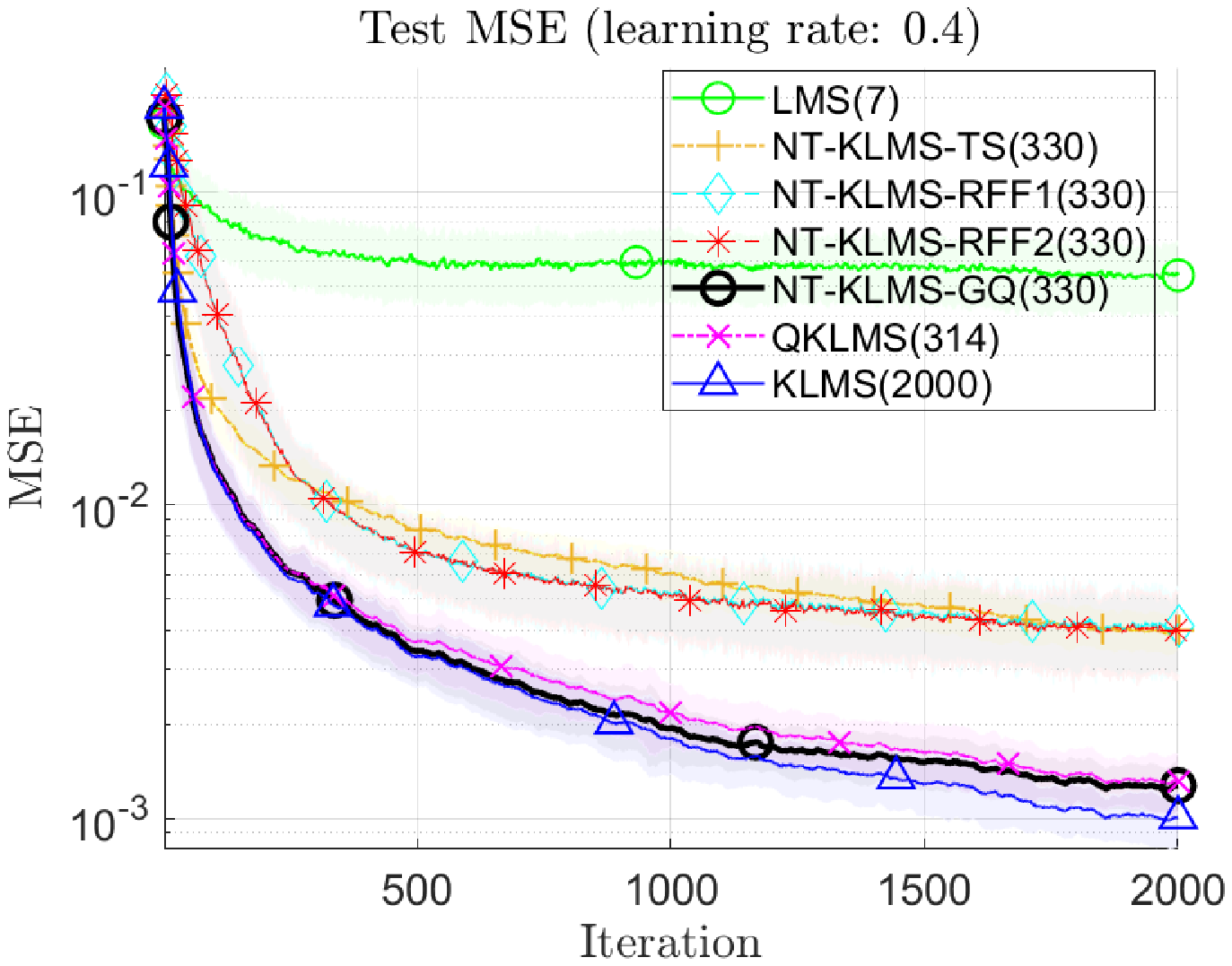}
\end{subfigure}
	\caption{Learning curves averaged over 200 independent runs. The input/feature dimension is listed for each adaptive LMS variant. Shaded regions correspond to $\pm 1 \sigma$.}
	\label{fig:MG_lr_performance}
\end{figure}

The data are standardized by subtracting its mean and dividing by its standard deviation, then followed by diving the resulting maximum absolute value to guarantee the sample values are within the range of $[-1,1]$ (to ensure the numerical error in approximation methods such as Taylor series expansion are manageable).  A time-embedding or input dimension of $d=7$ is used. The results are averaged over 200  independent trials. In each trial, 2000 consecutive samples with random starting point in the time series are used for training, and testing consists of 200 consecutive samples located in the future. 

In the first experiment, the feature space or RKHS dimension is set at $D = 330$, corresponding to Taylor polynomials of degrees up to 4, for input dimension $d = 7$. The GQ mapping is fixed across all trials, while RFFs are randomly generated in each trial. Taylor series expansion mappings are completely deterministic and fixed for all trials. We chose an $8$-th degree quadrature rule with sub-sampling to generate the explicit feature mapping. The learning curves for variants of the kernel least-mean-square algorithms are plotted in Fig. \ref{fig:MG_lr_performance} for three different learning rates ($\eta = 0.1, 0.2, 0.4$), with the linear LMS as a baseline. We see that a simple deterministic feature such as Taylor series expansion, can outperform random features, and the NT-KLMS GQ features outperformed all finite-dimensional RKHS filters, even the Quantized KLMS (QKLMS) with a comparable number of centers. The dictionary size of the QKLMS is impossible to fix \textit{a priori} across different trials, we fixed the vector quantization parameter ($q_{\rm factor} = 0.07$) such that the average value (314) of the dictionary size is close to the dimension of the finite-dimensional RKHS (330).

\begin{figure}[t]
	\centering
	\includegraphics[width=0.40\textwidth]{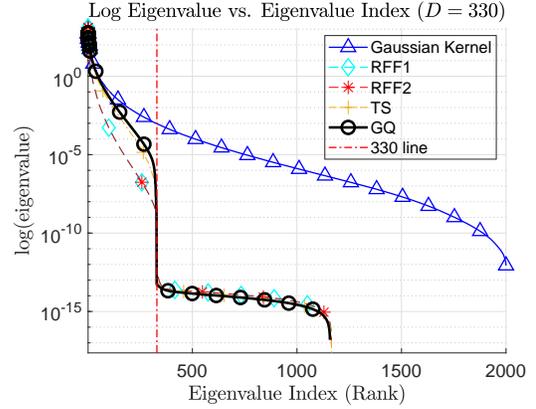}
	\caption{Eigenspectrum of the Gram matrices ($2000 \times 2000$ samples) computed using the Gaussian kernel and dot products of the feature maps, respectively, averaged over 200 independent trials.}
	\label{fig:Eigenspectrum}
\end{figure} 
\begin{figure*}[t]
	\centering
	\includegraphics[width=1\textwidth]{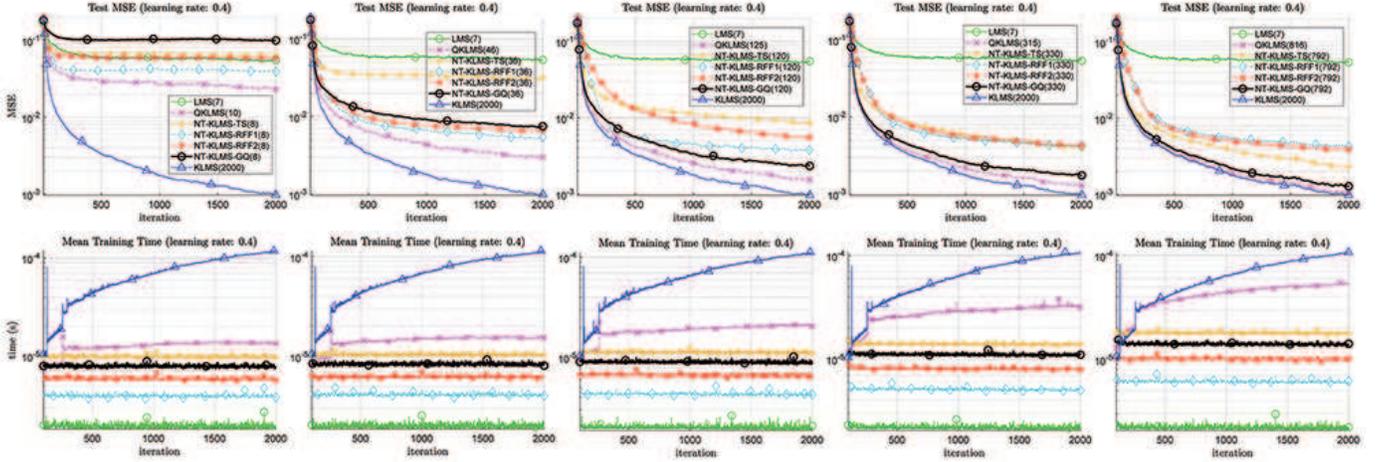}
	\caption{Learning curves and CPU times averaged over 200 independent runs. The input/feature dimension is listed for each adaptive RLS variant. Each column corresponds to an RKHS dimension determined by Taylor polynomials of incremental degree (starting from 1).}
	\label{fig:ComplexityVsPerformance}
\end{figure*} 
Next, we analyzed the eigenspectrum of the Gram matrices computed using the Gaussian kernel and the dot products of the feature maps, in Fig. \ref{fig:Eigenspectrum}. The eigenvalues of each Gram matrix ($2000\times2000$ samples) is plotted in descending order and averaged over 200 independent trials. Since the Gaussian kernel induces an infinite dimensional RKHS, we see that the nonnegative eigenspread is across the 2000 samples/dimensions, i.e., full rank for 2000 dimensions. The explicit feature maps induce a finite dimensional RKHS (dimension 330), so we see that the eigenvalues drop off to practically 0 after the 330th mark, i.e., rank 330 or the eigenvectors form an orthonormal set in the 330-D RKHS. From the rate of decay, we see that deterministic features such as GQ and TS fill the 330-D space much better, while the random Fourier features show wastage and can be well approximated using a much lower dimensional space.

\subsection{Complexity Analysis}
Furthermore, despite its popularity as one of the best sparsification algorithms that curb the growth of KAF, the complexity of QKLMS is not scalable like that of the finite-dimensional feature-space filter order, e.g., computing the inner product in the infinite-dimensional RKHS using the kernel trick requires $d\times |\mathcal{D}|$ operations, where $|\mathcal{D}|$ is the size of the dictionary, since for each input pair, the dot product is performed with respect to the input vector dimension $d$ (square norm argument in the Gaussian kernel), then repeated for all centers in the dictionary, not including the overhead to keep the dictionary size always in check. And, because the mapping is implicit and data-dependent for conventional kernel method, we can not compute this \emph{a priori} and must wait at the execution time. 

We illustrate this in Fig. \ref{fig:ComplexityVsPerformance}, performed on Intel Core i7-7700HQ using MATLAB. Again, the results are averaged over 200 independent trials. Here, for a more balanced comparison, the $8$-th degree GQ mappings are regenerated in each trail according to the sub-sampling distribution. Similarly, a different random Fourier mapping is redrawn. Again, the Taylor series mapping is constant and determined by the number of terms included in the expansion. In each column of Fig. \ref{fig:ComplexityVsPerformance}, we increment the number of TS terms included (corresponding to the maximum Taylor polynomial powers), starting from one, yielding $D = 8, 36, 120, 330, 792$ finite-dimensional RKHSs (for an input dimension of $d = 7$). To achieve a comparable dictionary size for QKLMS, we set the quantization parameters to $q_{\rm factor} = 1, 0.3, 0.14, 0.07, 0.026$, respectively. We see that GQ outperforms all other finite dimensional features for moderate size spaces (smaller number of features, in the first two columns, is insufficient to approximate the $8$-th degree quadrature rule). In this example, we see that even TS is better than random features at $D\geq 330$). To improve clarity, we plotted CPU time (seconds) in log scale. Clearly, finite-dimensional RKHS filtering or NT-KAF is much faster (constant time) and more scalable than unrestricted growth (linear with respect to the training data) and sparsification technique (linear with respect to the growing dictionary size plus sparsification overhead). We also show empirically that RFF1 (sine-cosine pair) is better than RFF2 (cosine with non-shift-invariant phase noise) for most cases and is faster (half the number of random features needed in each generation).   

\begin{figure}[t]
	\centering
	\begin{subfigure}
		\centering
		\includegraphics[width=0.40\textwidth]{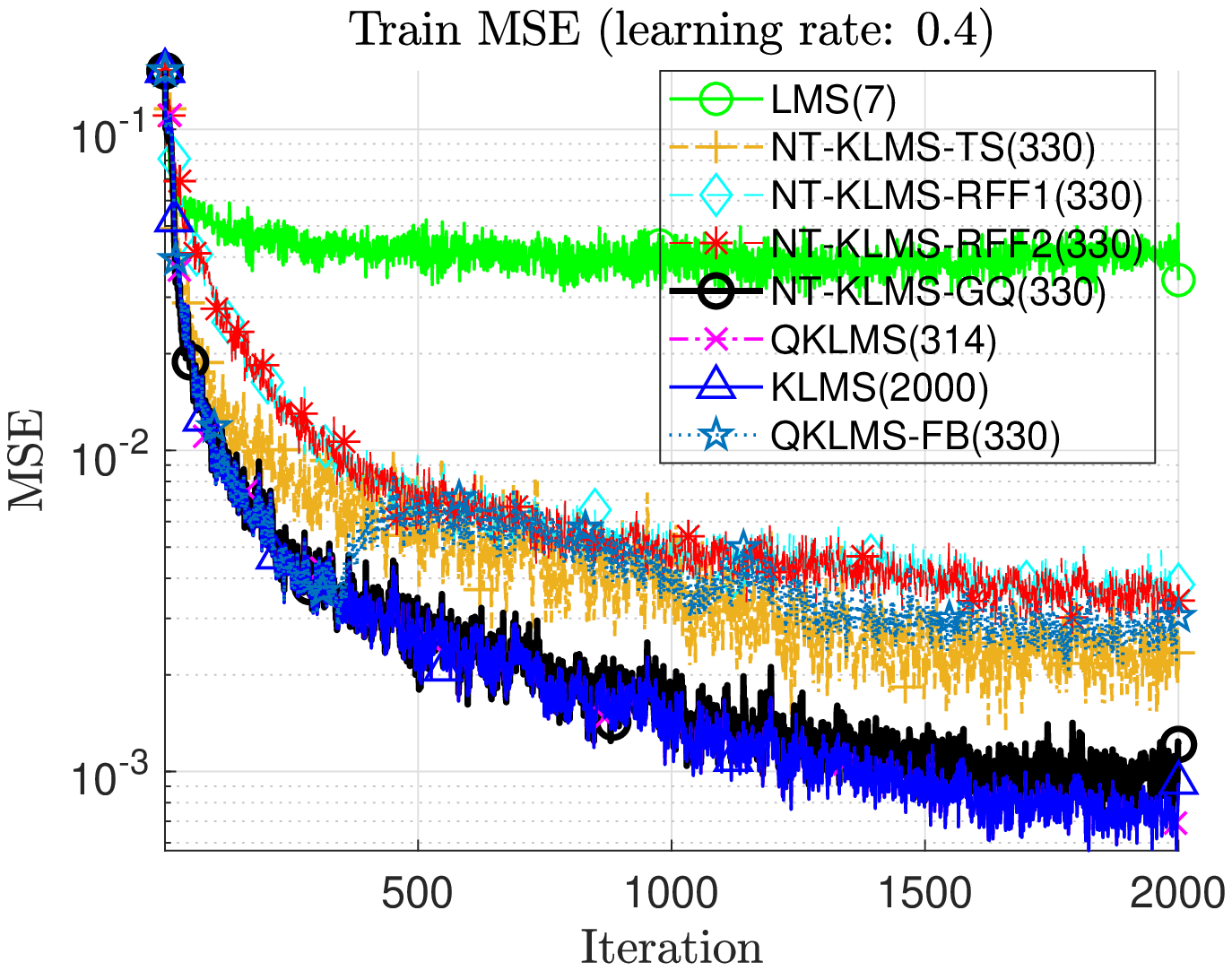}
	\end{subfigure}%
	\begin{subfigure}
		\centering
		\includegraphics[width=0.40\textwidth]{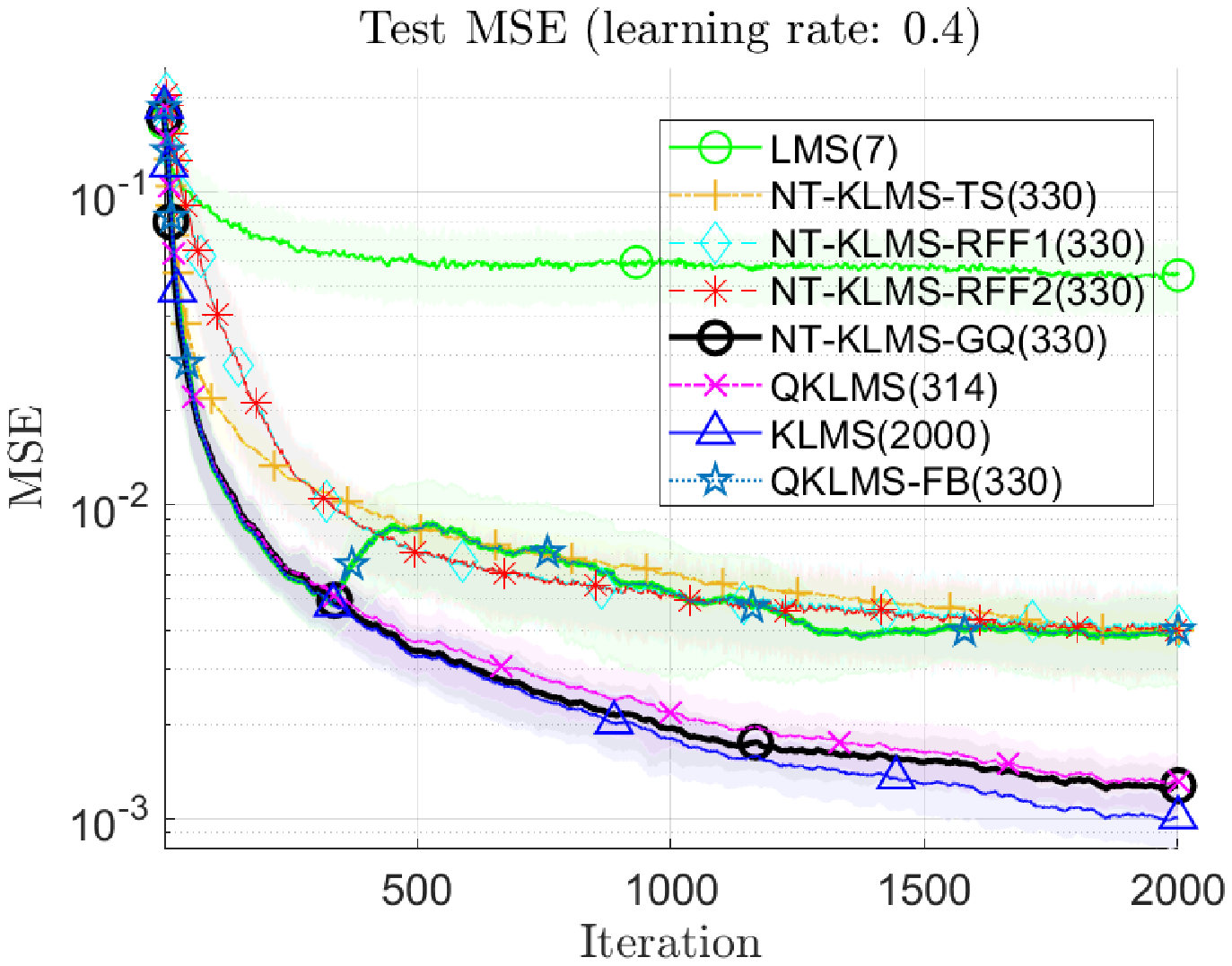}
	\end{subfigure}
	\begin{subfigure}
	\centering
	\includegraphics[width=0.40\textwidth]{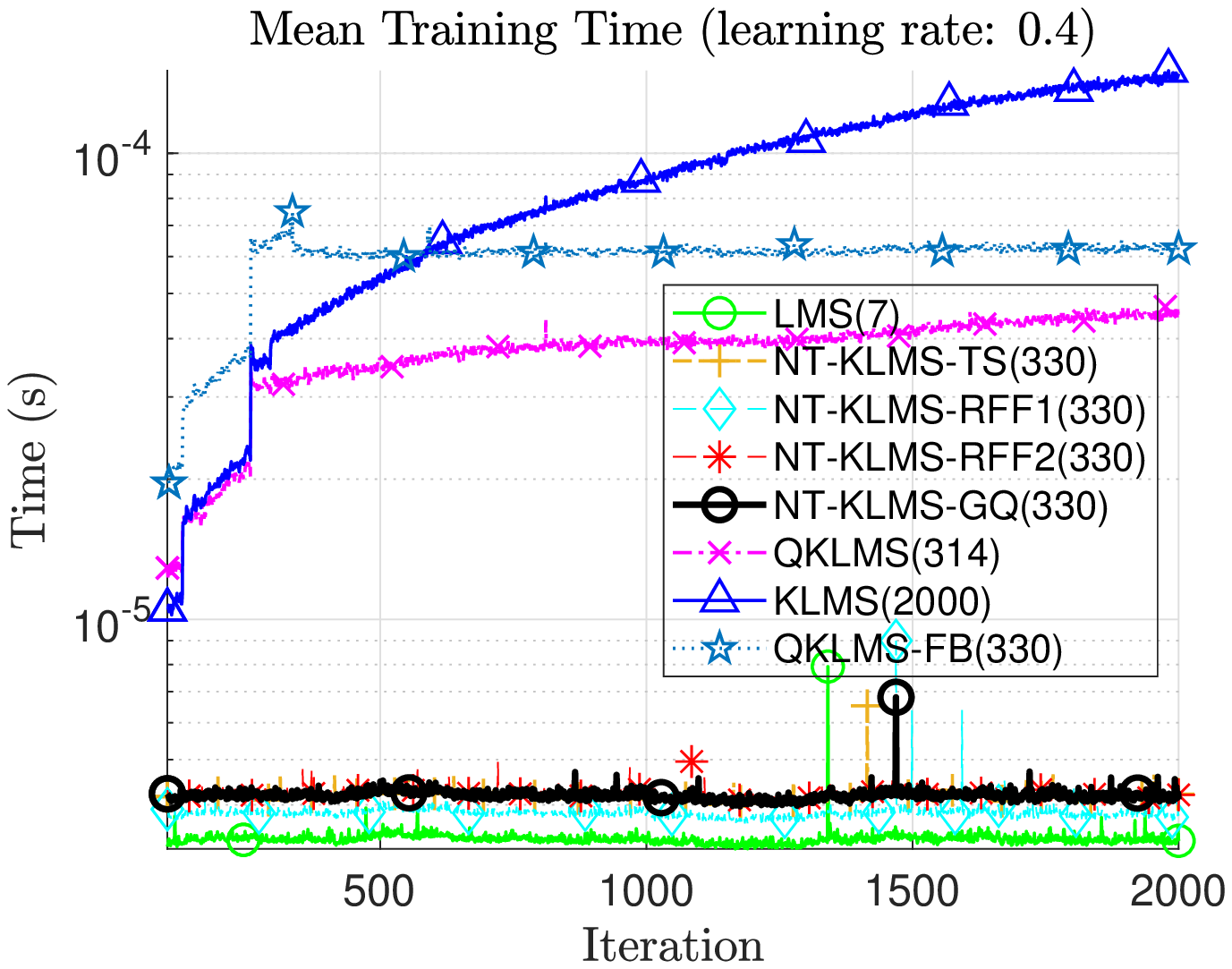}
\end{subfigure}
	\caption{Learning curves and CPU times averaged over 200 independent runs. FB-QKLMS quantization threshold of $q_{\rm factor} = 0.005$ and forgetting factor of $\lambda = 0.9$.}
	\label{fig:FB}
\end{figure}

As stated above, the dictionary size of QKLMS is impossible to set beforehand, to further show the superiority of our approach, we compared our method to a fixed dictionary formulation. Fixed-Budget (FB) QKLMS is an online filtering algorithm that discards centers using a significance measure to maintain a constant-sized dictionary. We compare its performance with the finite-dimensional RKHS no-trick kernel adaptive filtering in Fig. \ref{fig:FB}. We see that pruning techniques are not good for generalization, as the performances suffer from clipping after the first center is discarded.
\subsection{Noise Analysis}
\begin{figure}[t!]
	\centering
	\begin{subfigure}
		\centering
		\includegraphics[width=0.40\textwidth]{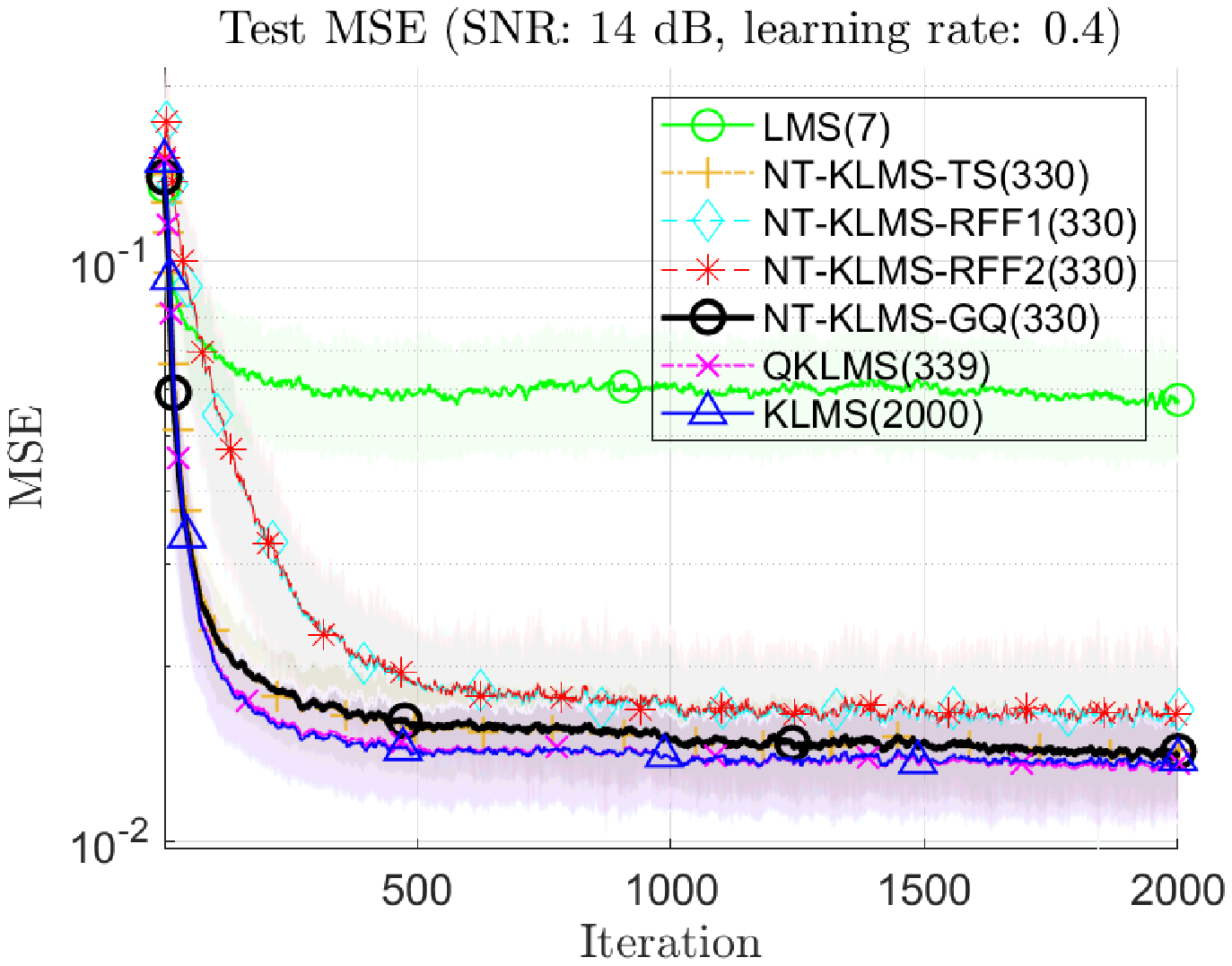}
	\end{subfigure}%
	\begin{subfigure}
		\centering
		\includegraphics[width=0.40\textwidth]{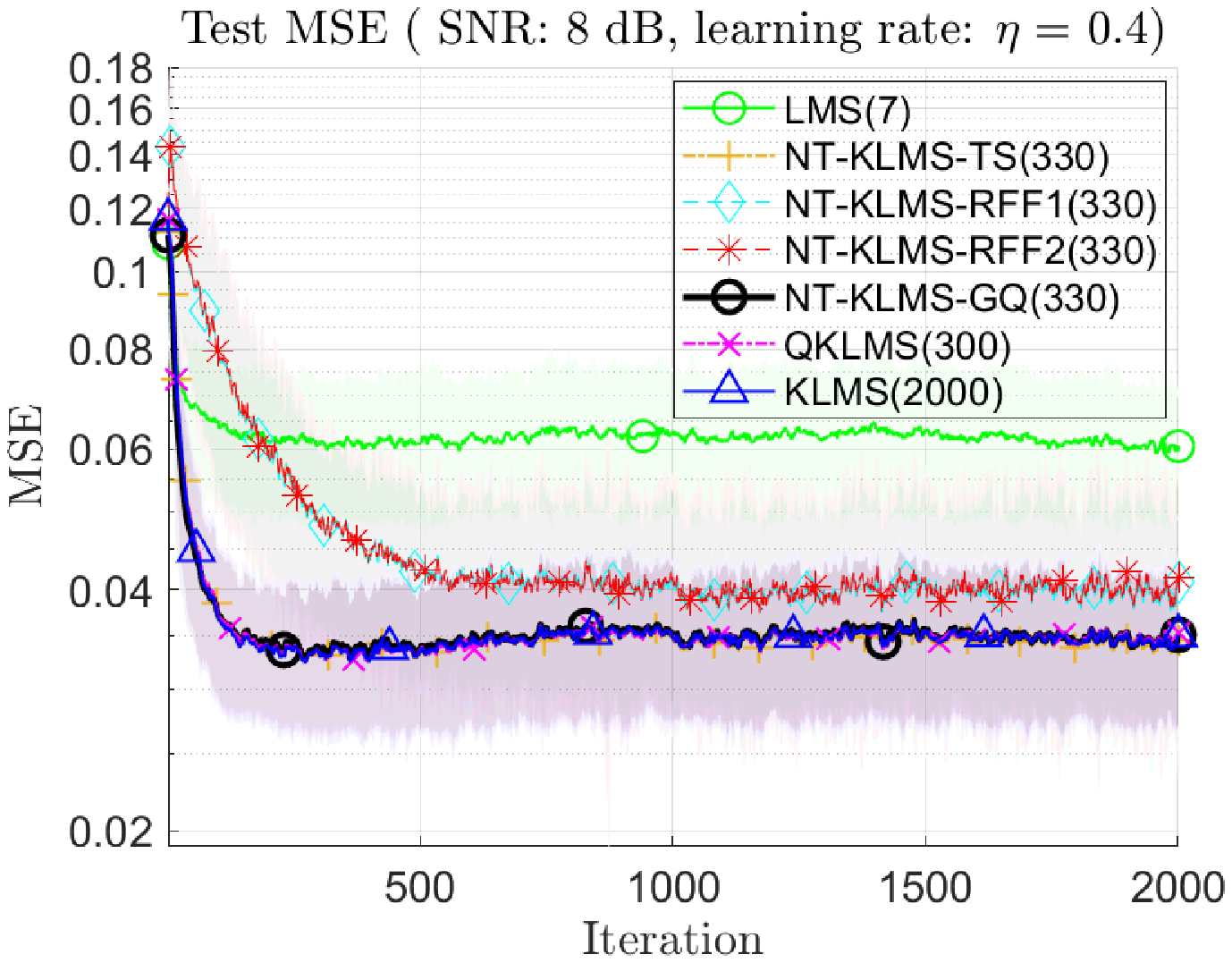}
	\end{subfigure}
	\caption{Learning curves averaged over 200 independent runs with signal-to-noise ratio (SNR) of 14 dB (top) and 8 dB (bottom). The input/feature dimension is listed for each adaptive LMS variant.}
	\label{fig:NoisePerformance}
\end{figure}
Next, we fixed the RKHS dimension to $D =330$ and evaluated the filter performance under noisy conditions by introducing additive white Gaussian noise (AWGN) to the dataset. The noise performances of KLMS variants are plotted in Fig. \ref{fig:NoisePerformance} and summarized in Table. \ref{tab:SNR}. We see that deterministic features outperformed random features, with very competitive results when compared to KLMS and QKLMS, especially given the space-time complexities of NT-KLMS. For low signal-to-noise ratio (SNR) settings, NT-KLMS can even outperform QKLMS and KLMS, e.g., SNR = 8 dB, demonstrating the robustness of no-trick kernel adaptive filtering.
\begin{table}[ht]\renewcommand{\arraystretch}{1.5}
	\tiny
	\centering\caption{Test Set MSE after 2000 training iterations (Feature dimension $D = 330$, learning rate $\eta = 0.4$).}
	\label{tab:SNR}
	\begin{tabular}{ |l|c|c|c|}
		\hline		
		\diagbox{\scriptsize Alg.}{\scriptsize SNR} & \small Clean  &\small 14 dB &\small 8 dB \\ \hline
	\scriptsize	LMS (7) & $0.0537 \pm 0.0113$ & $0.0575 \pm 0.0109$ & $0.0604 \pm 0.0107$\\ \hline
	\scriptsize	NT-KLMS-RFF1 & $0.0041 \pm 0.0012$ & $0.0168 \pm 0.0054$ & $0.0409 \pm 0.0163$\\ \hline
	\scriptsize	NT-KLMS-RFF2 & $0.0041 \pm 0.0012$ & $0.0171 \pm 0.0059$ & $0.0414 \pm 0.0179$\\ \hline
		\scriptsize	NT-KLMS-TS & $0.0039 \pm 0.0005$ & $0.0143 \pm 0.0029$ & $\mathbf{0.0346 \pm 0.0080}$\\ \hline
	\scriptsize	NT-KLMS-GQ & $\mathbf{0.0019 \pm 0.0003}$ & $\mathbf{0.0142 \pm 0.0028}$ & $0.0351 \pm 0.0080$\\ \hline\hline
	\scriptsize	QKLMS & $0.0012 \pm 0.0003$ & $0.0136 \pm 0.0028$ & $0.0353 \pm 0.0086$ \\ 
	\scriptsize (mean size) & (315) & (314) & (305)\\\hline
	\scriptsize	KLMS (2000) & $0.0010 \pm 0.0002$ & $0.0138 \pm 0.0029$ & $0.0350 \pm 0.0082$\\ \hline
	\end{tabular}
	\normalsize
\end{table}
\subsection{Chaotic Time Series Prediction using (K)RLS Algorithms} 
\begin{figure}[t]
	\centering
	\includegraphics[width=0.48\textwidth]{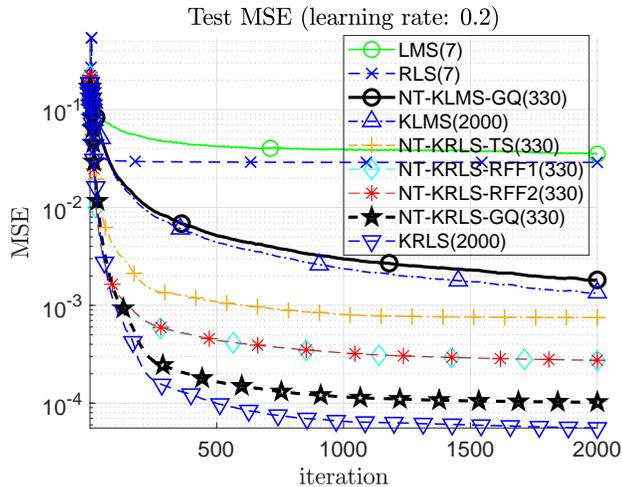}
	\caption{Test-set learning curves averaged over 200 independent runs. The input/feature dimension is listed for each adaptive LMS or RLS variant.}
	\label{fig:RLS}
\end{figure} 
Last but not least, the same principle can be easily applied to any linear filtering techniques to achieve nonlinear performance in a finite-dimension RKHS. Without loss of generality, we show the relative performances of recursive least square (RLS) based algorithms on the Mackey-Glass chaotic time series in Fig. \ref{fig:RLS}. The results are averaged over 200 independent trials, with input dimension of $d = 7$ and finite-dimensional RKHS of $D= 330$. Forgetting factor is set at $\lambda = 1$. The linear LMS and RLS, along with KLMS and KRLS, serve as baselines. As we can see, for stationary signal, RLS has extremely fast convergence and without mis-adjustment. Similarly, the kernelized version (KRLS) outperformed KLMS. The NT-KRLS using Gaussian quadrature features outperformed all other no-trick kernel adaptive filters.

\section{Conclusion}\label{Sec:Conclusion}
Kernel methods form a powerful, flexible, and theoretically-grounded unifying framework to solve nonlinear problems in signal processing and machine learning.	Since the explicit mapping is unavailable, the standard approach relies on the \emph{kernel trick} to perform pairwise evaluations of a kernel function, which leads to scalability issues for large datasets due to its linear and superlinear growth with respect to training data. A popular approach for handling this problem, known as random Fourier features, samples from a distribution to obtain the basis of a higher-dimensional feature space, where its inner product approximates the kernel function. In this paper, we proposed a different perspective by viewing these nonlinear features as defining an equivalent reproducing kernel in a new finite-dimensional reproducing kernel Hilbert space with universal approximation property. Furthermore, we proposed to use polynomial-exact decompositions to construction the feature mapping for kernel adaptive filtering. We demonstrated and validated our no-trick methodology to show that deterministic features are faster to generate and outperform state-of-the-art kernel approximation methods based on random features, and are significantly more efficient than conventional kernel methods using the \emph{kernel trick}.

In the future, we will extend no-trick kernel methods to information theoretic learning and advanced filtering algorithms.
\clearpage
	\bibliographystyle{IEEEtran}
	\bibliography{IEEEabrv,references}{}

 \newcommand{\noop}[1]{}
\begin{thebibliography}{10}
\providecommand{\url}[1]{#1}
\csname url@samestyle\endcsname
\providecommand{\newblock}{\relax}
\providecommand{\bibinfo}[2]{#2}
\providecommand{\BIBentrySTDinterwordspacing}{\spaceskip=0pt\relax}
\providecommand{\BIBentryALTinterwordstretchfactor}{4}
\providecommand{\BIBentryALTinterwordspacing}{\spaceskip=\fontdimen2\font plus
\BIBentryALTinterwordstretchfactor\fontdimen3\font minus
  \fontdimen4\font\relax}
\providecommand{\BIBforeignlanguage}[2]{{%
\expandafter\ifx\csname l@#1\endcsname\relax
\typeout{** WARNING: IEEEtran.bst: No hyphenation pattern has been}%
\typeout{** loaded for the language `#1'. Using the pattern for}%
\typeout{** the default language instead.}%
\else
\language=\csname l@#1\endcsname
\fi
#2}}
\providecommand{\BIBdecl}{\relax}
\BIBdecl

\bibitem{rahimi2007RFF}
\BIBentryALTinterwordspacing
A.~Rahimi and B.~Recht, ``Random features for large-scale kernel machines,'' in
  \emph{Proceedings of the 20th International Conference on Neural Information
  Processing Systems}, ser. NIPS'07.\hskip 1em plus 0.5em minus 0.4em\relax
  USA: Curran Associates Inc., 2007, pp. 1177--1184. [Online]. Available:
  \url{http://dl.acm.org/citation.cfm?id=2981562.2981710}
\BIBentrySTDinterwordspacing

\bibitem{NIPS2009_3628}
\BIBentryALTinterwordspacing
Y.~Cho and L.~K. Saul, ``Kernel methods for deep learning,'' in \emph{Advances
  in Neural Information Processing Systems 22}, Y.~Bengio, D.~Schuurmans, J.~D.
  Lafferty, C.~K.~I. Williams, and A.~Culotta, Eds.\hskip 1em plus 0.5em minus
  0.4em\relax Curran Associates, Inc., 2009, pp. 342--350. [Online]. Available:
  \url{http://papers.nips.cc/paper/3628-kernel-methods-for-deep-learning.pdf}
\BIBentrySTDinterwordspacing

\bibitem{Huang2014}
P.~{Huang}, H.~{Avron}, T.~N. {Sainath}, V.~{Sindhwani}, and B.~{Ramabhadran},
  ``Kernel methods match deep neural networks on timit,'' in \emph{2014 IEEE
  International Conference on Acoustics, Speech and Signal Processing
  (ICASSP)}, May 2014, pp. 205--209.

\bibitem{Wilson2016}
\BIBentryALTinterwordspacing
A.~G. Wilson, Z.~Hu, R.~Salakhutdinov, and E.~P. Xing, ``Stochastic variational
  deep kernel learning,'' in \emph{Proceedings of the 30th International
  Conference on Neural Information Processing Systems}, ser. NIPS'16.\hskip 1em
  plus 0.5em minus 0.4em\relax USA: Curran Associates Inc., 2016, pp.
  2594--2602. [Online]. Available:
  \url{http://dl.acm.org/citation.cfm?id=3157382.3157388}
\BIBentrySTDinterwordspacing

\bibitem{Li2018}
\BIBentryALTinterwordspacing
K.~Li and J.~C. Pr{\'{\i}}ncipe, ``Biologically-inspired spike-based automatic
  speech recognition of isolated digits over a reproducing kernel hilbert
  space,'' \emph{Frontiers in Neuroscience}, vol.~12, p. 194, 2018. [Online].
  Available: \url{https://www.frontiersin.org/article/10.3389/fnins.2018.00194}
\BIBentrySTDinterwordspacing

\bibitem{Dao2017}
\BIBentryALTinterwordspacing
T.~Dao, C.~D. Sa, and C.~R{\'e}, ``Gaussian quadrature for kernel features,''
  in \emph{Proceedings of the 31st International Conference on Neural
  Information Processing Systems}, ser. NIPS'17.\hskip 1em plus 0.5em minus
  0.4em\relax USA: Curran Associates Inc., 2017, pp. 6109--6119. [Online].
  Available: \url{http://dl.acm.org/citation.cfm?id=3295222.3295359}
\BIBentrySTDinterwordspacing

\bibitem{QKLMS}
B.~Chen, S.~Zhao, P.~Zhu, and J.~C. Pr{\'{\i}}ncipe, ``Quantized kernel least
  mean square algorithm,'' \emph{IEEE Trans. Neural Netw. Learn. Syst.},
  vol.~23, no.~1, pp. 22--32, 2012.

\bibitem{NICE}
K.~Li and J.~C. Príncipe, ``Transfer learning in adaptive filters: The nearest
  instance centroid-estimation kernel least-mean-square algorithm,'' \emph{IEEE
  Transactions on Signal Processing}, vol.~65, no.~24, pp. 6520--6535, Dec
  2017.

\bibitem{SNIPGOAL}
K.~{Li} and J.~C. {Príncipe}, ``Surprise-novelty information processing for
  gaussian online active learning (snip-goal),'' in \emph{2018 International
  Joint Conference on Neural Networks (IJCNN)}, July 2018, pp. 1--6.

\bibitem{Singh12}
A.~{Singh}, N.~{Ahuja}, and P.~{Moulin}, ``Online learning with kernels:
  Overcoming the growing sum problem,'' in \emph{2012 IEEE International
  Workshop on Machine Learning for Signal Processing}, Sep. 2012, pp. 1--6.

\bibitem{Qi15}
Y.~Qi, G.~T. Cinar, V.~M.~A. Souza, G.~E. A. P.~A. Batista, Y.~Wang, and J.~C.
  Principe, ``Effective insect recognition using a stacked autoencoder with
  maximum correntropy criterion,'' in \emph{Proc. IJCNN}, Killarney, Ireland,
  2015, pp. 1--7.

\bibitem{Bouboulis18}
P.~{Bouboulis}, S.~{Chouvardas}, and S.~{Theodoridis}, ``Online distributed
  learning over networks in rkh spaces using random fourier features,''
  \emph{IEEE Transactions on Signal Processing}, vol.~66, no.~7, pp.
  1920--1932, April 2018.

\bibitem{Sutherland2015}
\BIBentryALTinterwordspacing
D.~J. Sutherland and J.~Schneider, ``On the error of random fourier features,''
  in \emph{Proceedings of the Thirty-First Conference on Uncertainty in
  Artificial Intelligence}, ser. UAI'15.\hskip 1em plus 0.5em minus 0.4em\relax
  Arlington, Virginia, United States: AUAI Press, 2015, pp. 862--871. [Online].
  Available: \url{http://dl.acm.org/citation.cfm?id=3020847.3020936}
\BIBentrySTDinterwordspacing

\bibitem{Sun2018}
\BIBentryALTinterwordspacing
Y.~Sun, A.~Gilbert, and A.~Tewari, ``But how does it work in theory? linear svm
  with random features,'' in \emph{Proceedings of the 32Nd International
  Conference on Neural Information Processing Systems}, ser. NIPS'18.\hskip 1em
  plus 0.5em minus 0.4em\relax USA: Curran Associates Inc., 2018, pp.
  3383--3392. [Online]. Available:
  \url{http://dl.acm.org/citation.cfm?id=3327144.3327257}
\BIBentrySTDinterwordspacing

\bibitem{pmlr-v97-li19k}
\BIBentryALTinterwordspacing
Z.~Li, J.-F. Ton, D.~Oglic, and D.~Sejdinovic, ``Towards a unified analysis of
  random {F}ourier features,'' in \emph{Proceedings of the 36th International
  Conference on Machine Learning}, ser. Proceedings of Machine Learning
  Research, K.~Chaudhuri and R.~Salakhutdinov, Eds., vol.~97.\hskip 1em plus
  0.5em minus 0.4em\relax Long Beach, California, USA: PMLR, 09--15 Jun 2019,
  pp. 3905--3914. [Online]. Available:
  \url{http://proceedings.mlr.press/v97/li19k.html}
\BIBentrySTDinterwordspacing

\bibitem{cotter2011explicit}
A.~Cotter, J.~Keshet, and N.~Srebro, ``Explicit approximations of the gaussian
  kernel,'' 2011.

\bibitem{Greengard1991}
\BIBentryALTinterwordspacing
L.~Greengard and J.~Strain, ``The fast gauss transform,'' \emph{SIAM J. Sci.
  Stat. Comput.}, vol.~12, no.~1, pp. 79--94, Jan. 1991. [Online]. Available:
  \url{https://doi.org/10.1137/0912004}
\BIBentrySTDinterwordspacing

\bibitem{Yang2003}
{Yang}, {Duraiswami}, {Gumerov}, and {Davis}, ``Improved fast gauss transform
  and efficient kernel density estimation,'' in \emph{Proceedings Ninth IEEE
  International Conference on Computer Vision}, Oct 2003, pp. 664--671 vol.1.

\bibitem{Bochner1959}
\BIBentryALTinterwordspacing
S.~Bochner, M.~Functions, S.~Integrals, H.~Analysis, M.~Tenenbaum, and
  H.~Pollard, \emph{Lectures on Fourier Integrals. (AM-42)}.\hskip 1em plus
  0.5em minus 0.4em\relax Princeton University Press, 1959. [Online].
  Available: \url{http://www.jstor.org/stable/j.ctt1b9s09r}
\BIBentrySTDinterwordspacing

\bibitem{Smolyak63}
{S. A. Smolyak}, ``Quadrature and interpolation formulas for tensor products of
  certain classes of functions,'' \emph{Dokl. Akad. Nauk SSSR}, vol. 148,
  no.~5, pp. 1042--1045, 1963.

\bibitem{Scholkopf01}
B.~Scholkopf and A.~J. Smola, \emph{Learning with Kernels, Support Vector
  Machines, Regularization, Optimization and Beyond}.\hskip 1em plus 0.5em
  minus 0.4em\relax Cambridge, MA, USA: MIT Press, 2001.

\bibitem{Vapnik95}
V.~Vapnik, \emph{The Nature of Statistical Learning Theory}.\hskip 1em plus
  0.5em minus 0.4em\relax New York: Springer-Verlag, 1995.

\bibitem{Scholkopf98}
B.~Scholkopf, A.~J. Smola, and K.~Muller, ``Nonlinear component analysis as a
  kernel eigenvalue problem,'' \emph{Neural Comput.}, vol.~10, no.~5, Jul.
  1998.

\bibitem{Rasmussen06}
C.~E. Rasmussen and C.~K.~I. Williams, \emph{Gaussian Processes for Machine
  Learning}.\hskip 1em plus 0.5em minus 0.4em\relax Cambridge, MA, USA: the MIT
  Press, 2006.

\bibitem{Liu10}
W.~Liu, J.~C. Pr{\'{\i}}ncipe, and S.~Haykin, \emph{Kernel Adaptive Filtering:
  A Comprehensive Introduction}.\hskip 1em plus 0.5em minus 0.4em\relax
  Hoboken, NJ, USA: Wiley, 2010.

\bibitem{KAARMA}
K.~Li and J.~C. Pr{\'{\i}}ncipe, ``The kernel adaptive
  autoregressive-moving-average algorithm,'' \emph{IEEE Trans. Neural Netw.
  Learn. Syst.}, vol.~27, no.~2, pp. 334--346, Feb. 2016.

\bibitem{li2019functional}
K.~Li and J.~C. Principe, ``Functional bayesian filter,'' 2019.

\bibitem{KLMS}
W.~Liu, P.~Pokharel, and J.~C. Pr{\'{\i}}ncipe, ``The kernel least-mean-square
  algorithm,'' \emph{{IEEE} Trans. Signal Process.}, vol.~56, no.~2, pp.
  543--554, 2008.

\bibitem{RT}
B.~Scholkopf, R.~Herbrich, and A.~J. Smola, ``A generalized representer
  theorem,'' in \emph{Proc. 14th Annual Conf. Comput. Learning Theory}, 2001,
  pp. 416--426.

\bibitem{Mackey77}
M.~C. Mackey and L.~Glass, ``Oscillation and chaos in physiological control
  systems,'' \emph{Science}, vol. 197, no. 4300, pp. 287--289, Jul. 1977.

\end{thebibliography}
		\begin{IEEEbiography}
		[{\includegraphics[width=1in,height=1.25in,clip,keepaspectratio]{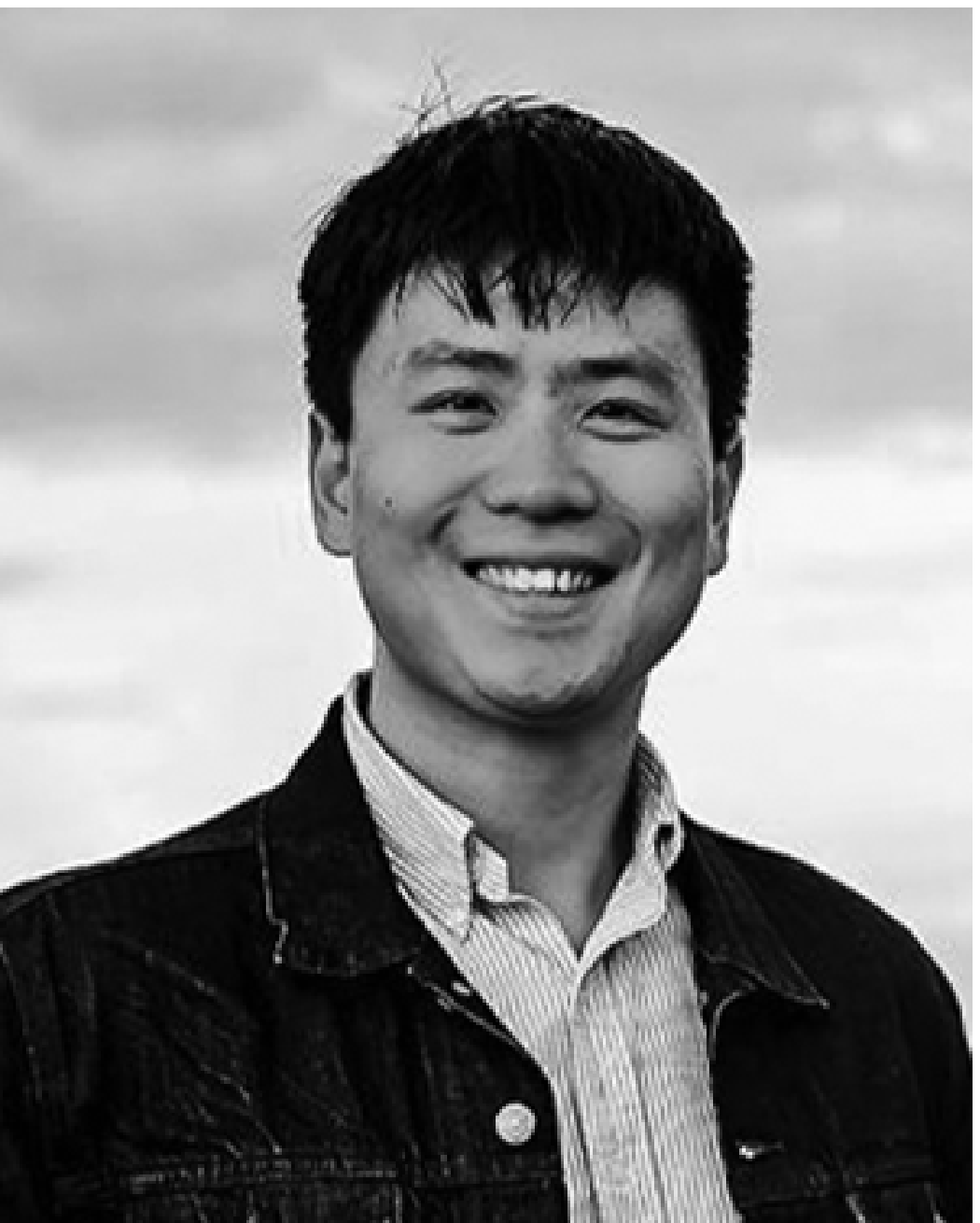}}]
		{Kan Li} (S'08) received the B.A.Sc. degree in electrical engineering from the University of Toronto in 2007, the M.S. degree in electrical engineering from the University of Hawaii in 2010, and the Ph.D. degree in electrical engineering from the University of Florida in 2015.  He is currently a research scientist at the University of Florida. His research interests include machine learning and signal processing.
	\end{IEEEbiography}
	
	\vspace{-8 mm}
	\begin{IEEEbiography}
		[{\includegraphics[width=1in,height=1.25in,clip,keepaspectratio]{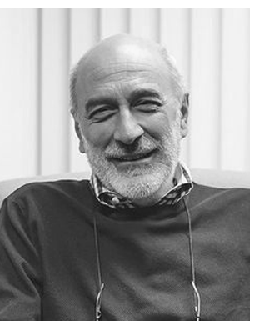}}]{Jos\'{e} C. Pr\'{i}ncipe}
		(M'83-SM'90-F'00) is the BellSouth and Distinguished Professor of Electrical and Biomedical Engineering at the University of Florida, and the Founding Director of the Computational NeuroEngineering Laboratory (CNEL). His primary research interests
		are in advanced signal processing with information theoretic criteria and adaptive models in reproducing kernel Hilbert spaces (RKHS), with application to brain-machine interfaces (BMIs). Dr.  Pr\'{i}ncipe is a Fellow of the IEEE, ABME, and AIBME. He is the past
		Editor in Chief of the IEEE Transactions on Biomedical Engineering, past Chair
		of the Technical Committee on Neural Networks of the IEEE Signal Processing
		Society, Past-President of the International Neural Network Society, and a recipient of the IEEE EMBS Career Award and the IEEE Neural Network Pioneer
		Award.
	\end{IEEEbiography}	
\end{document}